\documentclass{article}

\usepackage[T1]{fontenc}
\usepackage[utf8]{inputenc}
\usepackage{amsmath,amssymb,amsfonts,mathrsfs,bm}
\usepackage{mathtools}
\usepackage{amsthm}
\usepackage{scalerel}
\usepackage{nicefrac}
\usepackage{microtype} 
\usepackage[shortlabels]{enumitem}
\usepackage{graphicx}
\usepackage{epstopdf}
\DeclareGraphicsExtensions{.eps,.png,.jpg,.pdf}

\usepackage{url}
\usepackage{colortbl}
\usepackage{booktabs}
\usepackage{multirow}
\usepackage[normalem]{ulem}
\usepackage{xparse}
\usepackage{calc}
\usepackage{etoolbox}

\usepackage{array}
\newcolumntype{L}[1]{>{\raggedright\let\newline\\\arraybackslash\hspace{0pt}}m{#1}}
\newcolumntype{C}[1]{>{\centering\let\newline\\\arraybackslash\hspace{0pt}}m{#1}}
\newcolumntype{R}[1]{>{\raggedleft\let\newline\\\arraybackslash\hspace{0pt}}m{#1}}

\usepackage{glossaries}
\makeatletter
\sfcode`\.1006

\let\oldgls\gls
\let\oldglspl\glspl

\newcommand\fussy@ifnextchar[3]{%
	\let\reserved@d=#1%
	\def\reserved@a{#2}%
	\def\reserved@b{#3}%
	\futurelet\@let@token\fussy@ifnch}
\def\fussy@ifnch{%
	\ifx\@let@token\reserved@d
		\let\reserved@c\reserved@a
	\else
		\let\reserved@c\reserved@b
	\fi
	\reserved@c}

\renewcommand{\gls}[1]{%
\oldgls{#1}\fussy@ifnextchar.{\@checkperiod}{\@}}
\renewcommand{\glspl}[1]{%
\oldglspl{#1}\fussy@ifnextchar.{\@checkperiod}{\@}}

\newcommand{\@checkperiod}[1]{%
	\ifnum\sfcode`\.=\spacefactor\else#1\fi
}

\robustify{\gls}
\robustify{\glspl}
\makeatother

\newacronym{wrt}{w.r.t.}{with respect to}
\newacronym{RHS}{R.H.S.}{right-hand side}
\newacronym{LHS}{L.H.S.}{left-hand side}
\newacronym{iid}{i.i.d.}{independent and identically distributed}

\usepackage{hyperref}

\usepackage[capitalize,noabbrev]{cleveref}

\crefname{equation}{}{}
\Crefname{equation}{}{}
\crefname{claim}{claim}{claims}
\crefname{step}{step}{steps}
\crefname{line}{line}{lines}
\crefname{condition}{condition}{conditions}
\crefname{dmath}{}{}
\crefname{dseries}{}{}
\crefname{dgroup}{}{}

\crefname{Problem}{Problem}{Problems}
\crefformat{Problem}{Problem~(#2#1#3)}
\crefrangeformat{Problem}{Problems~(#3#1#4) to~(#5#2#6)}

\crefname{Theorem}{Theorem}{Theorems}
\crefname{Corollary}{Corollary}{Corollaries}
\crefname{Proposition}{Proposition}{Propositions}
\crefname{Lemma}{Lemma}{Lemmas}
\crefname{Definition}{Definition}{Definitions}
\crefname{Example}{Example}{Examples}
\crefname{Assumption}{Assumption}{Assumptions}
\crefname{Remark}{Remark}{Remarks}
\crefname{Rem}{Remark}{Remarks}
\crefname{remarks}{Remarks}{Remarks}
\crefname{Appendix}{Appendix}{Appendices}
\crefname{Supplement}{Supplement}{Supplements}
\crefname{Exercise}{Exercise}{Exercises}
\crefname{Theorem_A}{Theorem}{Theorems}
\crefname{Corollary_A}{Corollary}{Corollaries}
\crefname{Proposition_A}{Proposition}{Propositions}
\crefname{Lemma_A}{Lemma}{Lemmas}
\crefname{Definition_A}{Definition}{Definitions}

\usepackage{crossreftools}

\ifx\loadbreqn\undefined
	\relax
\else
	\usepackage{breqn}
\fi

\makeatletter
\def\cleartheorem#1{%
    \expandafter\let\csname#1\endcsname\relax
    \expandafter\let\csname c@#1\endcsname\relax
}
\def\clearthms#1{ \@for\tname:=#1\do{\cleartheorem\tname} }
\makeatother

\ifx\renewtheorem\undefined
	\ifx\useTheoremCounter\undefined
		\newtheorem{Theorem}{Theorem}
		\newtheorem{Corollary}{Corollary}
		\newtheorem{Proposition}{Proposition}
		
	\else
		\newtheorem{Theorem}{Theorem}
		\newtheorem{Corollary}[Theorem]{Corollary}
		
	\fi

	\newtheorem{Definition}{Definition}
	
	\newtheorem{Remark}{Remark}

\fi

\theoremstyle{remark}

\theoremstyle{plain}

\newcommand{\qednew}{\nobreak \ifvmode \relax \else
		\ifdim\lastskip<1.5em \hskip-\lastskip
			\hskip1.5em plus0em minus0.5em \fi \nobreak
		\vrule height0.75em width0.5em depth0.25em\fi}



\newcommand{\nn}{\nonumber\\ }

\NewDocumentCommand{\movedownsub}{e{^_}}{%
	\IfNoValueTF{#1}{%
		\IfNoValueF{#2}{^{}}
	}{%
		^{#1}
	}%
	\IfNoValueF{#2}{_{#2}}
}

\let\latexchi\chi
\RenewDocumentCommand{\chi}{}{\latexchi\movedownsub}

\newcommand{\Real}{\mathbb{R}}

\newcommand{\calC}{\mathcal{C}}

\newcommand{\calE}{\mathcal{E}}

\newcommand{\calG}{\mathcal{G}}

\newcommand{\calN}{\mathcal{N}}
\newcommand{\calO}{\mathcal{O}}

\newcommand{\calQ}{\mathcal{Q}}

\newcommand{\calV}{\mathcal{V}}

\newcommand{\ba}{\mathbf{a}}

\newcommand{\bc}{\mathbf{c}}

\newcommand{\bd}{\mathbf{d}}

\newcommand{\be}{\mathbf{e}}

\newcommand{\bi}{\mathbf{i}}

\newcommand{\bl}{\mathbf{l}}

\newcommand{\bn}{\mathbf{n}}

\newcommand{\bp}{\mathbf{p}}

\newcommand{\bq}{\mathbf{q}}

\newcommand{\br}{\mathbf{r}}

\newcommand{\bs}{\mathbf{s}}

\newcommand{\bt}{\mathbf{t}}

\DeclareSymbolFont{bsfletters}{OT1}{cmss}{bx}{n}
\DeclareSymbolFont{ssfletters}{OT1}{cmss}{m}{n}
\DeclareMathSymbol{\bsfGamma}{0}{bsfletters}{'000}
\DeclareMathSymbol{\ssfGamma}{0}{ssfletters}{'000}
\DeclareMathSymbol{\bsfDelta}{0}{bsfletters}{'001}
\DeclareMathSymbol{\ssfDelta}{0}{ssfletters}{'001}
\DeclareMathSymbol{\bsfTheta}{0}{bsfletters}{'002}
\DeclareMathSymbol{\ssfTheta}{0}{ssfletters}{'002}
\DeclareMathSymbol{\bsfLambda}{0}{bsfletters}{'003}
\DeclareMathSymbol{\ssfLambda}{0}{ssfletters}{'003}
\DeclareMathSymbol{\bsfXi}{0}{bsfletters}{'004}
\DeclareMathSymbol{\ssfXi}{0}{ssfletters}{'004}
\DeclareMathSymbol{\bsfPi}{0}{bsfletters}{'005}
\DeclareMathSymbol{\ssfPi}{0}{ssfletters}{'005}
\DeclareMathSymbol{\bsfSigma}{0}{bsfletters}{'006}
\DeclareMathSymbol{\ssfSigma}{0}{ssfletters}{'006}
\DeclareMathSymbol{\bsfUpsilon}{0}{bsfletters}{'007}
\DeclareMathSymbol{\ssfUpsilon}{0}{ssfletters}{'007}
\DeclareMathSymbol{\bsfPhi}{0}{bsfletters}{'010}
\DeclareMathSymbol{\ssfPhi}{0}{ssfletters}{'010}
\DeclareMathSymbol{\bsfPsi}{0}{bsfletters}{'011}
\DeclareMathSymbol{\ssfPsi}{0}{ssfletters}{'011}
\DeclareMathSymbol{\bsfOmega}{0}{bsfletters}{'012}
\DeclareMathSymbol{\ssfOmega}{0}{ssfletters}{'012}

\makeatletter
\newcommand*\rel@kern[1]{\kern#1\dimexpr\macc@kerna}
\newcommand*\widebar[1]{%
  \begingroup
  \def\mathaccent##1##2{%
    \rel@kern{0.8}%
    \overline{\rel@kern{-0.8}\macc@nucleus\rel@kern{0.2}}%
    \rel@kern{-0.2}%
  }%
  \macc@depth\@ne
  \let\math@bgroup\@empty \let\math@egroup\macc@set@skewchar
  \mathsurround\z@ \frozen@everymath{\mathgroup\macc@group\relax}%
  \macc@set@skewchar\relax
  \let\mathaccentV\macc@nested@a
  \macc@nested@a\relax111{#1}%
  \endgroup
}
\makeatother

\newcommand{\ifbcdot}[1]{\ifblank{#1}{\cdot}{#1}}

\DeclarePairedDelimiterX\abs[1]{\lvert}{\rvert}{\ifbcdot{#1}}
\DeclarePairedDelimiterX\parens[1]{(}{)}{\ifbcdot{#1}}
\DeclarePairedDelimiterX\brk[1]{[}{]}{\ifbcdot{#1}}
\DeclarePairedDelimiterX\braces[1]{\{}{\}}{\ifbcdot{#1}}
\DeclarePairedDelimiterX\angles[1]{\langle}{\rangle}{\ifblank{#1}{\cdot,\cdot}{#1}}
\DeclarePairedDelimiterX\ip[2]{\langle}{\rangle}{\ifbcdot{#1},\ifbcdot{#2}}
\DeclarePairedDelimiterX\norm[1]{\lVert}{\rVert}{\ifbcdot{#1}}
\DeclarePairedDelimiterX\ceil[1]{\lceil}{\rceil}{\ifbcdot{#1}}
\DeclarePairedDelimiterX\floor[1]{\lfloor}{\rfloor}{\ifbcdot{#1}}

\DeclareFontFamily{U}{matha}{\hyphenchar\font45}
\DeclareFontShape{U}{matha}{m}{n}{
      <5> <6> <7> <8> <9> <10> gen * matha
      <10.95> matha10 <12> <14.4> <17.28> <20.74> <24.88> matha12
      }{}
\DeclareSymbolFont{matha}{U}{matha}{m}{n}
\DeclareFontSubstitution{U}{matha}{m}{n}

\DeclareFontFamily{U}{mathx}{\hyphenchar\font45}
\DeclareFontShape{U}{mathx}{m}{n}{
      <5> <6> <7> <8> <9> <10>
      <10.95> <12> <14.4> <17.28> <20.74> <24.88>
      mathx10
      }{}
\DeclareSymbolFont{mathx}{U}{mathx}{m}{n}
\DeclareFontSubstitution{U}{mathx}{m}{n}

\DeclareMathDelimiter{\vvvert}{0}{matha}{"7E}{mathx}{"17}
\DeclarePairedDelimiterX\vertiii[1]{\vvvert}{\vvvert}{\ifbcdot{#1}}

\DeclarePairedDelimiterXPP\trace[1]{\operatorname{Tr}}{(}{)}{}{\ifbcdot{#1}} 
\DeclarePairedDelimiterXPP\col[1]{\operatorname{col}}{\{}{\}}{}{\ifbcdot{#1}} 
\DeclarePairedDelimiterXPP\row[1]{\operatorname{row}}{\{}{\}}{}{\ifbcdot{#1}} 
\DeclarePairedDelimiterXPP\erf[1]{\operatorname{erf}}{(}{)}{}{\ifbcdot{#1}}
\DeclarePairedDelimiterXPP\erfc[1]{\operatorname{erfc}}{(}{)}{}{\ifbcdot{#1}}
\DeclarePairedDelimiterXPP\KLD[2]{D}{(}{)}{}{\ifbcdot{#1}\, \delimsize\|\, \ifbcdot{#2}} 
\DeclarePairedDelimiterXPP\op[2]{\operatorname{#1}}{(}{)}{}{#2} 

\newcommand{\eqa}[1]{\stackrel{#1}{=}}
\newcommand{\ed}{\eqa{\mathrm{d}}}

\newcommand{\T}{^{\mkern-1.5mu\mathop\intercal}}
\newcommand{\ud}{\,\mathrm{d}} 

\DeclarePairedDelimiterXPP\indicate[1]{{\bf 1}}{\{}{\}}{}{\ifbcdot{#1}}

\providecommand\given{}

\DeclarePairedDelimiterX\Set[2]\{\}{%
\renewcommand\given{\SetSymbol[\delimsize]{#1}}
#2
}
\DeclarePairedDelimiterX\Setc[1]\{\}{%
\renewcommand\given{\SetSymbol{:}}
#1
}

\NewDocumentCommand\set{s o m}{%
	\IfBooleanTF#1%
	{\IfValueTF{#2}{\Set*{#2}{#3}}{\Setc*{#3}}}%
	{\IfValueTF{#2}{\Set{#2}{#3}}{\Setc{#3}}}%
}

\NewDocumentCommand{\evalat}{ s O{\big} m e{_^} }{%
\IfBooleanTF{#1}%
{\left. #3 \right|}{#3#2|}%
\IfValueT{#4}{_{#4}}%
\IfValueT{#5}{^{#5}}%
}

\providecommand\given{}
\DeclarePairedDelimiterXPP\cprob[1]{}(){}{
\renewcommand\given{\nonscript\,\delimsize\vert\allowbreak\nonscript\,\mathopen{}}%
#1%
}
\DeclarePairedDelimiterXPP\cexp[1]{}[]{}{
\renewcommand\given{\nonscript\,\delimsize\vert\allowbreak\nonscript\,\mathopen{}}%
#1%
}

\DeclareDocumentCommand \P { s e{_^} d() g } {%
	\mathbb{P}%
	\IfBooleanTF{#1}%
		{
			\IfValueT{#2}{_{#2}}%
			\IfValueT{#3}{^{#3}}%
			\IfValueTF{#5}{\cprob{#4 \given #5}}{\IfValueT{#4}{\cprob{#4}}}%
		}%
		{
			\IfValueT{#2}{_{#2}}%
			\IfValueT{#3}{^{#3}}%
			\IfValueTF{#5}{\cprob*{#4 \given #5}}{\IfValueT{#4}{\cprob*{#4}}}%
		}%
}

\DeclareDocumentCommand \E { s e{_^} o g } {%
	\mathbb{E}%
	\IfBooleanTF{#1}%
		{
			\IfValueT{#2}{_{#2}}%
			\IfValueT{#3}{^{#3}}%
			\IfValueTF{#5}{\cexp{#4 \given #5}}{\IfValueT{#4}{\cexp{#4}}}%
		}%
		{
			\IfValueT{#2}{_{#2}}%
			\IfValueT{#3}{^{#3}}%
			\IfValueTF{#5}{\cexp*{#4 \given #5}}{\IfValueT{#4}{\cexp*{#4}}}%
		}%
}

\ExplSyntaxOn
\NewDocumentCommand \dist {m o o} {%
\mathrm{#1}\left(%
	\IfValueT{#3}{%
		\tl_if_blank:nTF{ #3 }{\cdot\, \middle|\, }{#3\, \middle|\, }%
	}
	\IfValueT{#2}{#2}%
\right)%
}
\ExplSyntaxOff

\NewDocumentCommand {\cbrace} {t+ D[]{black} D(){\widthof{#5}} m m } {%
	\begingroup%
		\color{#2}
		\IfBooleanTF{#1}{%
			\overbrace{#4}^%
		}{
			\underbrace{#4}_%
		}%
		{\parbox[c]{#3}{\centering\footnotesize{#5}}}%
	\endgroup%
}

\let\oldforall\forall
\renewcommand{\forall}{\oldforall \, }

\let\oldexist\exists
\renewcommand{\exists}{\oldexist \, }

\graphicspath{{./Figures/}{./figures/}}
\DeclareDocumentCommand{\includeCroppedPdf}{ o O{./Figures/} m }{
	\IfFileExists{#2#3-crop.pdf}{}{%
		\immediate\write18{pdfcrop #2#3.pdf #2#3-crop.pdf}}%
	\includegraphics[#1]{#2#3-crop.pdf}
}

\makeatletter
\newcommand*{\addFileDependency}[1]{
  \typeout{(#1)}
  \@addtofilelist{#1}
  \IfFileExists{#1}{}{\typeout{No file #1.}}
}
\makeatother

\definecolor{gray90}{gray}{0.9}

\ifx\nohighlights\undefined
	\newcommand{\red}[1]{{\color{red} #1}}
	\newcommand{\blue}[1]{{{\color{blue} #1}}}

	\newcommand{\cyan}[1]{{\color{cyan} #1}}
	\newcommand{\msout}[1]{\text{\color{green} \sout{\ensuremath{#1}}}}
	
	\newcommand{\del}[1]{{\color{green}\ifmmode \msout{#1}\else\sout{#1}\fi}}
\else
	\newcommand{\red}[1]{#1}
	\newcommand{\blue}[1]{#1}

	\newcommand{\cyan}[1]{#1}
	\newcommand{\msout}[1]{#1}
	\newcommand{\del}[1]{#1}
\fi

\newcommand{\hhide}[1]{}

\ifx\diagnoselabel\undefined
	\relax
\else
	\makeatletter
	\def\@testdef #1#2#3{%
		\def\reserved@a{#3}\expandafter \ifx \csname #1@#2\endcsname
			\reserved@a  \else
			\typeout{^^Jlabel #2 changed:^^J%
				\meaning\reserved@a^^J%
				\expandafter\meaning\csname #1@#2\endcsname^^J}%
			\@tempswatrue \fi}
	\makeatother
\fi

\newcommand{\bit}[1]{\textbf{\textit{#1}}}
\newcommand{\tb}[1]{\textbf{#1}}

\def\p{\partial}

\DeclareMathOperator{\id}{id}

\newcommand{\tvb}[3]{\left(\frac{\partial}{\partial {#1}^{#2}}\right)_{\negmedspace #3}}

\DeclareFontFamily{U}{MnSymbolC}{}
\DeclareSymbolFont{MnSyC}{U}{MnSymbolC}{m}{n}
\DeclareMathSymbol{\diamondplus}{\mathbin}{MnSyC}{"7C}
\DeclareMathSymbol{\diamonddot}{\mathbin}{MnSyC}{"7E}
\DeclareFontShape{U}{MnSymbolC}{m}{n}{
    <-6>  MnSymbolC5
   <6-7>  MnSymbolC6
   <7-8>  MnSymbolC7
   <8-9>  MnSymbolC8
   <9-10> MnSymbolC9
  <10-12> MnSymbolC10
  <12->   MnSymbolC12}{}

\def\id{\text{id}}

\def\p{\partial}

\def\se{\subseteq}

\usepackage[accepted]{icml2023}

\usepackage{bbding}
\usepackage{tikz}
\usepackage{tikz,pgfplots}
\pgfplotsset{compat=1.18}

\usepackage{tikz-cd}
\DeclareMathOperator{\st}{s.t.\ }
\usetikzlibrary{shapes.geometric, arrows, patterns,calc}
\usepackage{IEEEtrantools}

\theoremstyle{plain}

\theoremstyle{definition}

\theoremstyle{remark}

\usepackage[textsize=tiny]{todonotes}

\newcommand{\first}[1]{\red{\tb{#1}}}
\newcommand{\second}[1]{\blue{\tb{#1}}}
\newcommand{\third}[1]{\cyan{\tb{#1}}}

\newcommand\smt[1]{\calC^{\infty}({#1})}

\def\bcon{\begin{convention}} 
\def\econ{\end{convention}}
\def\bc{\begin{cora}}
\def\ec{\end{cora}}
\def\bcl{\begin{rema}}
\def\ecl{\end{rema}}
\def\bd{\begin{defa}}
\def\ed{\end{defa}}
\def\ben{\begin{enumerate}}
\def\benr{\begin{enumerate}[label=(\roman*)]}
\def\een{\end{enumerate}}
\def\be{\begin{align}}
\def\ee{\end{align}}
\def\bse{\begin{align*}}
\def\ese{\end{align*}}
\def\bex{\begin{exma}}
\def\eex{\end{exma}}
\def\bexe{\begin{exercise}}
\def\eexe{\end{exercise}}
\def\bit{\begin{itemize}}
\def\eit{\end{itemize}}
\def\bl{\begin{lema}}
\def\el{\end{lema}}
\def\bnn{\begin{rema}}
\def\enn{\end{rema}}
\def\bn{\begin{note}}
\def\en{\end{note}}
\def\bp{\begin{lema}}
\def\ep{\end{lema}}
\def\bpo{\begin{postulate}}
\def\epo{\end{postulate}}
\def\bq{\begin{proof}}
\def\eq{\end{proof}}
\def\br{\begin{rema}}
\def\er{\end{rema}}
\def\bs{\begin{solution}}
\def\es{\end{solution}}
\def\btab{\begin{table}}
\def\etab{\end{table}}
\def\btb{\begin{tabular}}
\def\etb{\end{tabular}}
\def\bter{\begin{defa}}
\def\eter{\end{defa}}
\def\bt{\begin{thma}}
\def\et{\end{thma}}
\def\btik{\begin{tikzpicture}}
\def\etik{\end{tikzpicture}}
  \def\bi{\begin{IEEEeqnarray*}}
  \def\ei{\end{IEEEeqnarray*}}
    \def\ba{\begin{array}}
  \def\ea{\end{array}}

\def\b{\beta}
\def\del{\delta}

\def\d{\mathrm{d}}

\def\b1{\mathbb{1}} 

\def\R{\mathbb{R}} 

\def\bbox{\begin{tcolorbox}[colback=white!5,colframe=blue!75!black,title=Exercise]}
\def\ebox{\end{tcolorbox}}

\newcommand{\cl}{:\text{ }}

 \def\bse{\begin{equation*}}
  \def\ese{\end{equation*}}
   \def\btab{\begin{table}}
  \def\etab{\end{table}}
  \def\btb{\begin{tabular}}
  \def\etb{\end{tabular}}

\begin{document}

\twocolumn[
\icmltitle{Node Embedding from Neural Hamiltonian Orbits in Graph Neural Networks}



\icmlsetsymbol{equal}{*}

\begin{icmlauthorlist}
\icmlauthor{Qiyu Kang}{equal,yyy}
\icmlauthor{Kai Zhao}{equal,yyy}
\icmlauthor{Yang Song}{comp}
\icmlauthor{Sijie Wang}{yyy}
\icmlauthor{Wee Peng Tay}{yyy}
\end{icmlauthorlist}

\icmlaffiliation{yyy}{School of Electrical and Electronic Engineering, Nanyang Technological University, Singapore}
\icmlaffiliation{comp}{C3 AI, Singapore}

\icmlcorrespondingauthor{Qiyu Kang}{kang0080@e.ntu.edu.sg}

\icmlkeywords{Machine Learning, ICML}

\vskip 0.3in
]



\printAffiliationsAndNotice{\icmlEqualContribution} 

\begin{abstract}
In the graph node embedding problem, embedding spaces can vary significantly for different data types, leading to the need for different GNN model types. In this paper, we model the embedding update of a node feature as a Hamiltonian orbit over time. Since the Hamiltonian orbits generalize the exponential maps, this approach allows us to learn the underlying manifold of the graph in training, in contrast to most of the existing literature that assumes a fixed graph embedding manifold with a closed exponential map solution. 
Our proposed node embedding strategy can automatically learn, without extensive tuning, the underlying geometry of any given graph dataset even if it has diverse geometries. 
We test Hamiltonian functions of different forms and verify the performance of our approach on two graph node embedding downstream tasks: node classification and link prediction.
Numerical experiments demonstrate that our approach adapts better to different types of graph datasets than popular state-of-the-art graph node embedding GNNs. The code is available at \url{https://github.com/zknus/Hamiltonian-GNN}.
\end{abstract}
\section{Introduction}
Graph neural networks (GNNs) \cite{yueBio2019, AshoorNC2020, kipf2017semi, ZhangTKDE2022, WuTNNLS2021} have shown remarkable inference performance on graph-structured data, including, but not limited to, social media networks, citation networks, and molecular graphs in chemistry.
Most existing GNNs embed graph nodes in Euclidean spaces without further consideration of the dataset graph geometry. For some graph structures like the tree-like graphs \cite{liu2019hyperbolic_GNN}, the Euclidean space may not be a proper choice for the node embedding. Recently, hyperbolic GNNs \cite{chami2019hyperbolic_GCNN, liu2019hyperbolic_GNN} propose to embed nodes into a hyperbolic space instead of the conventional Euclidean space. It has been shown that tree-like graphs can be inferred more accurately by hyperbolic GNNs.

\begin{figure}[!tb]
    \centering
\includegraphics[width=0.45\textwidth,trim={1cm 0 1.6cm 1cm},clip]{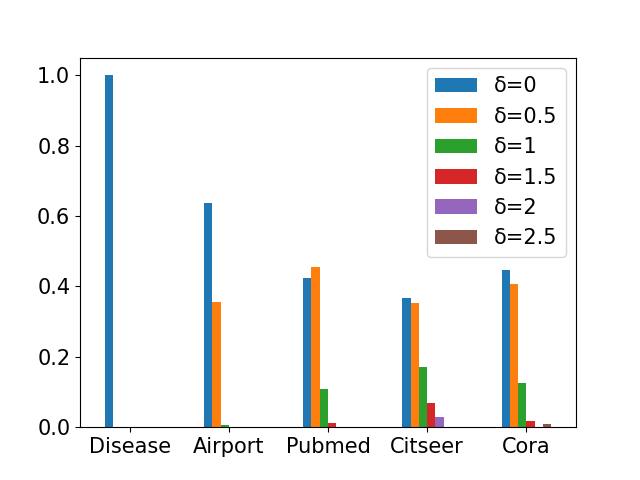}
    \caption{Gromov $\delta$-hyperbolicity distribution of datasets.}
    \label{fig:hyper}
    \vspace{-0.5cm}
\end{figure}

Real-world graphs often have complex and varied structures, which can be more effectively represented by utilizing different geometric spaces. As shown in \cref{fig:hyper}, the Gromov $\delta$-hyperbolicity distributions\footnote{If the $\delta$-hyperbolicity distribution is more concentrated at lower values, the more hyperbolic the graph dataset.} \cite{Gromov1987} of various datasets exhibit a range of diverse values. This indicates that it is not optimal to embed each dataset with diverse geometries into a single globally homogeneous geometry.
Works like  \cite{zhu2020GIL} have attempted to embed graph nodes in a mixture of the  Euclidean and hyperbolic spaces, where the intrinsic graph local geometry is attained from the mixing weight. In several studies, including \cite{gu2018learning, bachmann2020constant_curvature, lou2020differentiating}, researchers use (products of) constant curvature Riemannian spaces for graph node embedding where the spaces are assumed to be spherical, hyperbolic, or Euclidean. The work \cite{xiong2022pseudo} considers a special pseudo-Riemannian manifold named pseudo-hyperboloid that is of constant nonzero curvature and is diffeomorphic to the product manifolds of a unit sphere and the Euclidean space.

Embedding nodes in the aforementioned \emph{restricted} (pseudo-)Riemannian manifolds is achieved through the exponential map in closed forms, which is essentially a geodesic curve on the manifolds as the projected curve of the \emph{cogeodesic orbits} on the manifolds' cotangent bundles \cite{lee2013smooth,klingenberg2011riemannian}. In our work, we propose to embed the nodes, via more general \emph{Hamiltonian orbits}, into a general manifold, which generalizes the above graph node embedding works.

Manifolds have a diverse set of applications in physics, and their usage and development can be found interwoven throughout the literature. From the physics perspective, the cotangent bundles are the natural phase spaces in classical mechanics \cite{de2011generalized} where the physical system evolves according to the basic laws of physics modeled as differential equations on the phase spaces. In this paper, we propose a new GNN paradigm based on Hamiltonian mechanics \cite{Goldstein2001} with flexible Hamiltonian functions. 
Our objective is to design a new node embedding strategy that can automatically learn, without extensive tuning, the underlying geometry of any given graph dataset even if it has diverse geometries.
We enable the node features to evolve on the manifold under the influence of neighbors. 
The learnable Hamiltonian function on the manifold guides the node embedding evolution to follow a learnable law analogous to basic physical laws.

\textbf{Main contributions.}
Our main contributions are summarized as follows:
\begin{enumerate}[leftmargin=0.5cm,topsep=0pt,itemsep=-1ex,partopsep=1ex,parsep=1ex]
\item We consider the graph node embedding problem on an underlying manifold and enable node embedding through a learnable Hamiltonian orbit associated with the Hamiltonian scalar function on the manifold's cotangent bundle. 
\item Our node embedding strategy can automatically learn, without extensive tuning, the underlying
geometry of any given graph dataset even if it has diverse geometries. We empirically demonstrate its ability by testing on two graph node embedding downstream tasks: node classification and link prediction.
\item From empirical experiments, we observe that the oversmoothing problem of GNNs can be mitigated if the node features evolve through Hamiltonian orbits. By the conservative nature of the Hamiltonian equations, our model enables a stable training and inference process while updating the node features over time and layers.
\end{enumerate}
\section{Related Work}\label{sec:rel_wor}
While our paper is related to Hamiltonian neural networks in the literature, we are the first, to our best knowledge, to model graph node embedding with Hamiltonian equations. We briefly review Hamiltonian neural networks, Riemannian manifold GNNs, and physics-inspired GNNs in \cref{appsec:related}.

\emph{Notations:} We use the \emph{Einstein summation convention} \cite{lee2013smooth} for expressions with tensor indices. When using this convention, if an index variable appears twice in a term, once as a superscript and once as a subscript, it means a summation of the term over all possible values of the index variable. For example, $a^i b_i \triangleq \sum_{i=1}^d a^i b_i$. 

\section{Motivations and Preliminaries}\label{sec:pre}
In this section, we briefly review the concepts of the geodesic curve on a (pseudo-)Riemannian manifold from the principle of stationary action in the form of Lagrange’s equations. We then further generalize the geodesic curve to the Hamiltonian orbit associated with an energy function $H$, which is a conserved quantity along the orbit.
Our primary goal is to develop a more flexible and robust method for graph node embedding by leveraging the concepts of geodesic curves and Hamiltonian orbits on manifolds. We first summarize the motivation of our work as follows.

\tb{Motivation I: from the exponential map to the Riemannian geodesic.} The geodesic curve gives rise to the exponential map that maps points from the tangent space to the manifold and has been utilized in  \cite{chami2019hyperbolic_GCNN,bachmann2020constant_curvature,xiong2022pseudo} to enable graph node embedding in some restricted (pseudo-)Riemannian manifolds where closed forms of the exponential map are obtainable. From this perspective, by using the geodesic curve, we generalize the graph node embedding to an arbitrary pseudo-Riemannian manifold with {learnable local geometry $g$} using Lagrange’s equations. 

\tb{Motivation II: from geodesic to Hamiltonian orbit.} Despite the above conceptual generalization for node embedding using geodesic curves, the specific curve formulation involving minimization of curve length may result in a loss of generality for node feature evolution along the curve. We thus further generalize the geodesic curve to the Hamiltonian orbit associated with a more general energy function $H$ that is conserved along the orbit. In \cref{sec:frame}, we propose graph node embedding without an explicit metric by using Hamiltonian orbits with {learnable energy functions $H$}.

We kindly refer readers to \cref{sec.mot} for a more comprehensive elucidation that may enhance their understanding after we introduce the related concepts in \cref{sec.geo_exp}.

\subsection{Manifold and Riemannian Metric}
\tb{Manifold and local chart representation.} 
On a $d$-dimensional manifold $M$, for each point on $M$, there exists a triple $\{q, U, V\}$, called a chart, such that $U$ is an open neighborhood of the point in $M$, $V$ is an open subset of $\mathbb{R}^d$, and $q: U \rightarrow V$ is a homeomorphism, which gives us a coordinate representation for a local area in $M$. \\
\tb{Tangent and cotangent vector spaces.} For any point $q$ on $M$ (we identify each point covered by a local chart on $M$ by its representation $q$), we may assign two vector spaces named the tangent vector space $T_qM$ and cotangent vector space $T^*_qM$.  The vectors from the tangent and cotangent spaces can be interpreted as representing the velocity and generalized momentum, respectively, of an object's movement in classical mechanics.\\
\tb{Riemannian metric.} A Riemannian manifold is a manifold $M$ equipped with a \emph{Riemannian metric} $g$, where we assign to any point $q \in M$ and pair of vectors $u, v \in T_{q} M$ an \emph{inner product} $\langle{u}, {v}\rangle_{g(q)}.$
This assignment is assumed to be smooth with respect to the base point $q \in M$. The length of a tangent vector $u \in T_{q} M$ is then defined as
\begin{align}
\|u\|_{g(q)}:=\langle u, u\rangle_{g(q)}^{1 / 2} .
\end{align}
\begin{itemize}[leftmargin=0.5cm,topsep=0pt,itemsep=-1ex,partopsep=1ex,parsep=1ex]
    \item \tb{Local coordinates representation:} In local coordinates with $q=\left(q^1, \ldots, q^d\right)\T \in M, u=\left(u^1, \ldots, u^d\right)\T \in T_{q} M$ and $v=\left(v^1, \ldots, v^d\right)\T \in T_{q} M$, the Riemannian metric $g=g(q)$ is a real symmetric \emph{positive definite} matrix and the inner product above is given by
\begin{align}
\langle u, v\rangle_{g(q)}:= g_{i j}(q) u^i v^j
\end{align}
\item \tb{Pseudo-Riemannian metric:} We may generalize the Riemannian metric to a metric tensor that only requires a \emph{non-degenerate} condition \cite{lee2018introduction} instead of the stringent positive definiteness condition in the inner product. One example of a pseudo-Riemannian manifold is the Lorentzian manifold, which is important in applications of general relativity.
\end{itemize}

\subsection{Geodesic Curves and Exponential Maps} \label{sec.geo_exp}
\tb{Length and energy of a curve.} Let $q:[a, b] \rightarrow M$ be a smooth curve.\footnote{We abuse notations in denoting the chart coordinate map as $q$ and the curve as $q(t)$. It will be clear from the context which one is being referred to.} We use $\dot{q}$ and $\ddot{q}$ to denote the first and second order derivatives of $q(t)$, respectively. We define the following:

$\bullet$ length of the curve:
\begin{align}
\ell(q):=\int_a^b\|\dot{q}(t)\|_{g(q(t))} \ud t . \label{eq:len}
\end{align}
$\bullet$ energy of the curve:
\begin{align}
E(q):=\frac{1}{2} \int_a^b\|\dot{q}(t)\|_{g(q(t))}^2 \ud t . \label{eq:ene}
\end{align}
\tb{Geodesic curves.} On a Riemannian manifold, geodesic curves are defined as curves that have a minimal length as given by \cref{eq:len} and with two fixed endpoints $q(a)$ and $q(b)$.
However, computations based on minimizing the length to obtain the curves are difficult. It turns out that the minimizers of $E(q)$ also minimize $\ell(q)$  \cite{malham2016introduction}. Consequently, the geodesic curve formulation may be obtained by minimizing the energy of a smooth curve on $M$.\\
\tb{Principle of stationary action and Euler–Lagrange equation.} The Lagrangian function $L( q(t), \dot{q}(t))$ minimizes the following functional (in physics, the functional is known as an \tb{action})
\begin{align}
S(q)=\int_a^b L( q(t), \dot{q}(t)) \d t \label{eq:sq}
\end{align}
with two fixed endpoints at $t=a$ and $t=b$ only if  the following \emph{Euler–Lagrange equation} is satisfied:
\begin{align}
\frac{\partial L}{\partial q^i}(q(t), \dot{q}(t))- \frac{\d}{\d t} \frac{\partial L}{\partial \dot{q}^i}(q(t), \dot{q}(t))=0. \label{eq:euler_lag}
\end{align}
\tb{Geodesic equation for geodesic curves.}  The Euler–Lagrange equation derived from minimizing the energy \cref{eq:ene} with local coordinates representation,  
\begin{align}
    L=\frac{1}{2}\|\dot{q}(t)\|_{g(q(t))}^2 = \frac{1}{2}g_{i k}(q) \dot{q}^i \dot{q}^k \label{eq:Lgeo}
\end{align}
 is expressed as the following ordinary differential equations called the \emph{geodesic equation}:
\begin{align}
\ddot{q}^i+ \Gamma_{j k}^i \dot{q}^j \dot{q}^k=0, \label{eq:geoeq}
\end{align}
for all $i=1,\ldots,d$, where the Christoffel symbols 
$\Gamma_{j k}^i\coloneqq \frac{1}{2} g^{i \ell}\left(\frac{\partial g_{\ell j}}{\partial q_k}+\frac{\partial g_{k \ell}}{\partial q_j}-\frac{\partial g_{j k}}{\partial q_{\ell}}\right)$
and $[g^{ij}]$ denotes the inverse matrix of the matrix $[g_{ij}]$.
The solutions to the geodesic equation \cref{eq:geoeq} give us the geodesic curves.\\
\tb{Exponential map.} Given the geodesic curves, at each point $q\in M$, for velocity vector $v \in T_qM$, the \emph{exponential map} is defined to obtain the point on $M$ reached by the unique geodesic that passes through $q$ with velocity $v$ at time $t=1$ \cite{lee2018introduction}. Formally, we have 
\begin{align}
    \exp_q(v) = \gamma(1)
\end{align}
where  $\gamma(t)$ is the curve given by the geodesic equation \cref{eq:geoeq} with initial conditions $q(0)=q$ and $\dot{q}(0)=v$. 

With regards to \tb{Motivation I}, we note that \cite{chami2019hyperbolic_GCNN} considers graph node embedding over a homogeneous negative-curvature Riemannian manifold called hyperboloid manifold. In contrast, we generalize the embedding of nodes to an arbitrary pseudo-Riemannian manifold through the geodesic equation \cref{eq:geoeq} with a {learnable} metric $g$ that derives the local graph geometry from the nodes and their neighbors.

\subsection{From Geodesics to General Hamiltonian Orbits}\label{ssec:pre_ham_orb}
The geodesic curves and the derived exponential map essentially come from  \cref{eq:sq} with $L$ in \cref{eq:Lgeo} specified from the curve energy \cref{eq:ene}. 
However, the curves derived from this specific action may potentially sacrifice efficacy for the graph node embedding task since we do not know what a reasonable action formulation that guides the evolution of the node feature in this task is. 
Therefore, we follow the principle of stationary action but consider a {learnable action} that is more flexible than the length or energy of the curve. To better model the conserved quantity during the feature evolution, we reformulate the Lagrange equation to the Hamilton equation. This is our \tb{Motivation II}.\\
\tb{Hamiltonian function and equations.} The Hamiltonian orbit $(q(t), p(t))$ is given by the following \emph{Hamiltonian equations} with a \emph{Hamiltonian function} $H$:
\begin{align}
\dot{q}^i =\frac{\partial H}{\partial p_i}, \quad
\dot{p}_i =-\frac{\partial H}{\partial q^i}, \label{appeq:poincare1}
\end{align}
where $q$ is the local chart coordinate on the manifold while $p$ can be interpreted as a vector of {generalized momenta} in the cotangent vector space. In classical mechanics, the $2d$-dimensional pair $(q,p)$ is called \emph{phase space} coordinates that fully specify the state of a dynamic system with $p$ guiding the movement direction and speed. Later, we consider the node feature evolution following the trajectory specified by the  phase space coordinates.\\
\tb{Hamiltonian function vs. Lagrangian function.} The Hamiltonian function can be taken as the Legendre transform of the Lagrangian function:
\begin{align}
\begin{aligned}
      H(q,p)= p_i \dot{q}^i-L({q}, \dot{q}) \\
    \text{with } \dot{q}=\dot{q}(p) \text{ s.t. } p=\frac{\partial L}{\partial \dot{q}}. \label{eq:Legendre}
\end{aligned}
\end{align}
If $H$ is restricted to strictly convex functions, the Hamiltonian formalism is equivalent to a Lagrangian formalism \cite{de2011generalized}. \\
\tb{Geodesic equation reformulated as Hamiltonian function.} If $H$ is set as 
\begin{align}
    H(q,p)= \frac{1}{2}g^{i j}(q) p_i p_j, \label{eq:dae}
\end{align} 
where $[g^{ij}]$ denotes the inverse matrix of the matrix $[g_{ij}]$, we have the following Hamiltonian equations:
  \begin{align}
\begin{aligned}
\dot{q}^i =g^{i j} p_j, \quad
\dot{p}_i = -\frac{1}{2} \p_i g^{j k} p_j p_k. \label{eq:_geo}
\end{aligned}
\end{align}
The Hamiltonian orbit $(q(t),p(t))$, as the solution of \cref{eq:_geo}, gives us again the \emph{geodesic curves} $q(t)$ on the manifold $M$ if we only look at the first $d$-dimensional coordinates.
\begin{Theorem}[Conservation of energy \cite{da2008lectures}]\label{thm:energy_per}
    $H(q(t),p(t))$  is constant along the Hamiltonian orbit as solutions of \cref{appeq:poincare1}.
\end{Theorem}
In physics,  $H$ typically represents the total energy of a system, and \cref{thm:energy_per} indicates that the time evolution of the system follows the law of conservation of energy.
\section{Proposed Framework}\label{sec:frame}
We consider an undirected graph $\calG = (\calV, \calE)$ consisting of a finite set $\calV$ of vertices, together with a subset $\calE \subset \calV \times \calV$ of edges. Our objective is to learn a mapping $f$ that maps node features to embedding vectors:
\begin{align*}
    f\left(\mathcal{V}, \mathcal{E}\right) = Z \in \mathbb{R}^{|\mathcal{V}| \times d}. 
\end{align*}
The embedding $Z$ is supposed to capture both semantic and topological information. Since the input node features in most datasets are sparse, a fully connected (FC) layer is first applied to compress the raw input node features. 
Let ${}^nq$ be the $d$-dimensional compressed node feature for node $n$ after the FC layer.\footnote{We put the node index to the left of the variable to distinguish it from the manifold dimension index.} 
However, empirical experiments (see ``MLP'' results in \cref{sec:per}) indicate that for the graph node embedding task, such simple raw compressing without any consideration of the graph topology does not produce a good embedding.  Further graph neural network layers are thus required to update the node embedding. 

We consider the node features $\{{}^nq\}_{n\in\calV}$ to be located on an embedding manifold $M$ and take the node features as the chart coordinate representations for points on the manifold.
In \textbf{Motivations I and II} in \cref{sec:pre}, we have provided the rationale {for generalizing graph node embedding from the exponential map to the pseudo-Riemannian geodesic, and further to the Hamiltonian orbit.} To enforce the graph node feature update on the manifold with well-adapted learnable local geometry, we make use of the concepts from \cref{sec:pre}. 

\begin{figure*}[!htb]
    \centering
    \includegraphics[width=0.9\textwidth]{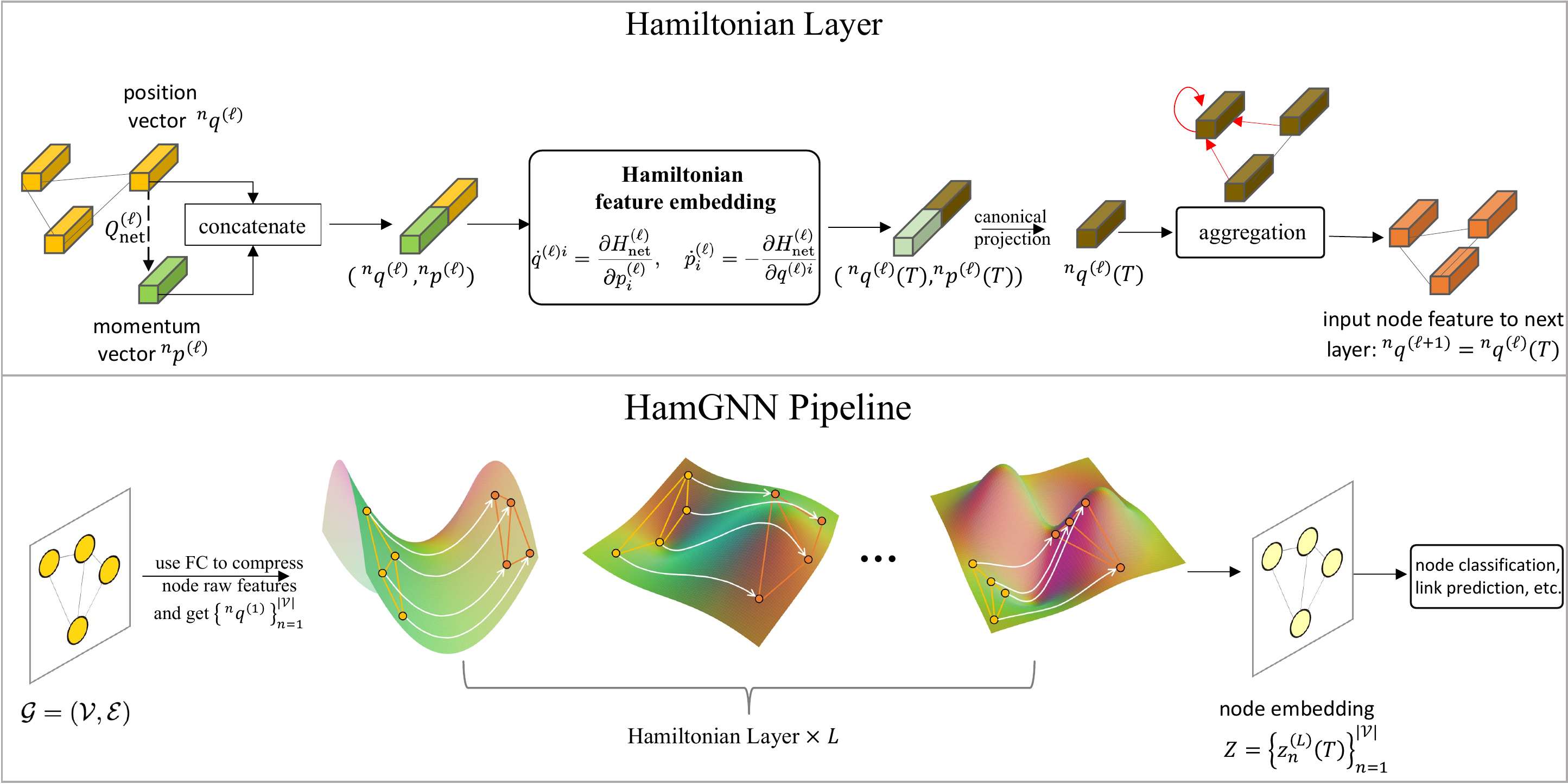}
   \caption{HamGNN architecture: in each layer, each node is assigned a learnable  ``momentum'' vector ${}^np$ (cf.\ \cref{eq:mom}) at time $t=0$, which initializes the evolution of the node feature. The node features evolve on a manifold following \cref{eq:ham_orb} to $({}^nq(T),{}^np(T))$ at the time $t=T$. We only take ${}^nq(T)$ as the embedding and input it to the next layer. After $L$ layers, we take ${}^nq^{(L)}(T)$ as the final node embedding.}
   \label{fig:model_arch}
\end{figure*}

\subsection{Model Architecture}\label{ssec:mod_arc}
\tb{Node feature evolution along Hamiltonian orbits in a Hamiltonian layer.} 
As introduced in \cref{ssec:pre_ham_orb}, the $2d$-dimensional \emph{phase space} coordinates $(q,p)$ fully specify a system's state. 
Consequently, for the node feature as a point on the manifold $M$, we associate to each point ${}^nq$ a \emph{learnable momentum vector ${}^np$}. This vector guides the direction and speed of the feature update, allowing the node feature to evolve along Hamiltonian orbits on the manifold. More specifically, we set 
\begin{align}
    {}^np = Q_{\mathrm{net}}({}^nq) \label{eq:mom}
\end{align}
where $Q_{\mathrm{net}}$\footnote{The subscript ``$\mathrm{net}$'' of a function $F_{\mathrm{net}}$ indicates that the function $F$ is parameterized by a neural network.} is instantiated by an FC layer. We consider a learnable Hamiltonian function $H_{\mathrm{net}}$ that specifies the node feature evolution trajectory in the phase space via the \emph{Hamiltonian equations}:
\begin{align}
\dot{q}^i =\frac{\partial  H_{\mathrm{net}}}{\partial p_i}, \quad
\dot{p}_i =-\frac{\partial  H_{\mathrm{net}}}{\partial q^i} \label{eq:ham_orb} 
\end{align}
with learnable Hamiltonian energy function
\begin{align}
    H_{\mathrm{net}}: (q,p)\mapsto \R.
\end{align}
The node features are updated along the Hamiltonian orbits, which are curves starting from each node $({}^nq,{}^np)$ at $t=0$. In other words, they are the solution of \cref{eq:ham_orb}
with the initial conditions $({}^nq(0),{}^np(0))=({}^nq,{}^np)$ at $t=0$. The solution of \cref{eq:ham_orb} on the phase space for each node $n\in\calV$ at time $T$ is given by a differential equation solver \cite{chen2018neural}, and denoted by $(^nq(T), ^np(T))$. The canonical projection $\pi({}^nq(T),{}^np(T))={}^nq(T)$ is taken to obtain the node feature embeddings on the manifold at time $T$. The aforementioned operations are performed within one layer, and we call it the \emph{Hamiltonian layer}. \\
\tb{Neighborhood Aggregation.}
After the node features update along the Hamiltonian orbits, we perform neighborhood aggregation on the features $\{{}^nq^{(\ell)}(T)\}_{n\in\calV}$, where $\ell$ indicates the $\ell$-th layer. Let $\mathcal{N}(n)=\{m:(n, m) \in \mathcal{E}\}$ denote the set of neighbors of node $n \in \mathcal{V}$. We only perform a simple yet efficient aggregation (see \cref{sec:exp}) for node $n$ as follows:
\begin{align}
{}^n q^{(\ell+1)}&={}^nq^{(\ell)}(T)+\frac{1}{|\calN(n)|}\sum_{m \in \mathcal{N}(n)}  {}^mq^{(\ell)}(T). \label{eq.agg}
\end{align}
\tb{Layer stacking for local geometry learning.}  
We stack up multiple Hamiltonian layers with neighborhood aggregation in between them. We first give an intuitive explanation for the case where $H_{\mathrm{net}}$ is set as \cref{eq:dae}. {A learnable metric $g_{\mathrm{net}}$} for the manifold is involved (see \cref{sec:H_geo} for more details) and the features are evolved following the geodesic curves with minimal length (see \cref{sec:pre}). 
Within each Hamiltonian layer, the metric $g_{\mathrm{net}}$ that is instantiated by a smooth FC layer only depends on the local node position on a pseudo-Riemannian manifold that varies from point to point.
Note that with layer stacking, these features contain information aggregated from their neighbors. 
The metric $g_{\mathrm{net}}$, therefore, learns from the graph topology, and each node is embedded with a local geometry that depends on its neighbors. 
In contrast, \cite{chami2019hyperbolic_GCNN,bachmann2020constant_curvature,xiong2022pseudo} consider graph node embedding using geodesic curves over some fixed manifold without adjustment of the local geometry. \\
At the beginning of \cref{sec:frame}, we have assumed the node features $\{{}^nq\}_n$ to be located in local charts of a preliminary embedding manifold $M$. The basic philosophy is that the embedding manifold evolves with a metric structure that adapts successively with neighborhood aggregation along multiple layers, whereas each node's features evolve to the most appropriate embedding on the manifold along the curves.
For a general learnable $H_{\mathrm{net}}$, the Hamiltonian orbit that starts from one node has aggregated information from its neighbors, which guides the learning of the curve that the node will be evolved along. 
Therefore, each node is embedded into a manifold with adaptation to the underlying geometry of any given graph dataset even if it has diverse geometries. \\
\tb{Conservation of $H_{\mathrm{net}}$.} From \cref{thm:energy_per}, the feature updating through the orbit indicates that the $H_{\mathrm{net}}$ is conserved along the curve. \\
\tb{Model summary.} Our model is called \emph{HamGNN} as we use Hamiltonian orbits for node feature updating on the manifold.  We summarize the HamGNN model architecture in \cref{fig:model_arch} and \cref{alg:hamgnn}. The forms of the Hamiltonian function $H_{\mathrm{net}}$ are given in  \cref{sec:diff_ham}.

\subsection{Different Hamiltonian Orbits}\label{sec:diff_ham}
We next propose different forms for $H_{\mathrm{net}}$ from which the corresponding Hamiltonian orbit and its variations are obtained in our GNN model. The node features are updated along the Hamiltonian orbits, which are curves starting from each node $({}^nq,{}^np)$ at $t=0$.  
Beginning with Motivation I from the paper, we design a learnable metric $g_{\mathrm{net}}$ in \cref{eq:dae} in \cref{sec:H_geo}. This approach relaxes the curve formulation constraint used in \cite{chami2019hyperbolic_GCNN,bachmann2020constant_curvature,xiong2022pseudo} and enables learnable geodesic curves to guide feature evolution on arbitrary (pseudo-)Riemannian manifolds while learning local geometry from graph datasets. To further extend learnable geodesic curves to learnable Hamiltonian orbits on manifolds, as stated in Motivation II, we introduce a flexible $H$ instantiated by an FC layer without constraints in Section \cref{sec:H_1}. Subsequently, we present variations based on \cref{sec:H_1} in Sections \ref{sec:H_2} through \ref{sec:H8} to examine their performance. In \cref{sec:H_2}, we add constraints to ensure that $H$ is a convex function, which allows us to equivalently test a more restricted Lagrangian formalism. In \cref{sec:H_4}, we include less restricted Hamiltonian mechanics without strict constant energy constraints. Finally, in \cref{sec:H8}, we consider a more flexible representation using the symplectic 2-form in comparison to \cref{sec:H_1}.

\subsubsection{Learnable Metric $g_{\mathrm{net}}$}\label{sec:H_geo}
In this subsection, we consider node embedding onto a pseudo-Riemannian and set $H_{\mathrm{net}}$  as \cref{eq:dae} where {a learnable metric $g_{\mathrm{net}}$} for the manifold is involved.
Within each Hamiltonian layer, the metric $g_{\mathrm{net}}$ instantiated by a smooth FC layer depends on the local node position on the pseudo-Riemannian manifold that varies from point to point. 
The output of $g_{\mathrm{net}}$ at position $q$ represents the inverse metric local representation $[g^{ij}]$.
However, from \cref{eq:dae}, the space complexity is order $d^3$ due to the partial derivative of $g$'s output being a $d\times d$ matrix. We therefore only consider \emph{diagonal metrics} to mitigate the space complexity. More specifically, we now define
\begin{align}
    g_{\mathrm{net}}(q)=\mathrm{diag}([\underbrace{-1,\ldots,-1}_{r},\underbrace{1,\ldots,1}_{s}]\odot h_{\mathrm{net}}(q))
\end{align}
where $h_{\mathrm{net}} : \Real^d\to\Real^d$ consists of non-linear trainable layers and  $\odot$ denotes element-wise multiplication.  To ensure non-degeneracy of the metric, the output of $h_{\mathrm{net}}$ is set to be away from $0$ with the final activation function of it being {strictly positive}. The vector $[-1,\ldots,-1,1,\ldots,1]$ controls the signature $ (r, s)$ \cite{lee2018introduction} of the metric $g$ with $r+s=d$, where $r$ and $s$ are the number of $-1$s and $1$s, respectively. The signature of the metric is set to be a hyperparameter. According to \cref{eq:_geo}, we have 
\begin{align}
\begin{aligned}
\dot{q}^i =g_{\mathrm{net}}^{i j} p_j, \quad
\dot{p}_i = -\frac{1}{2} \p_i g_{\mathrm{net}}^{j k} p_j p_k. \label{eq:H_geo}
\end{aligned}
\end{align}
 Intuitively, the node features evolve through the ``shortest'' curves on the manifold. The exponential map used in  \cite{chami2019hyperbolic_GCNN,bachmann2020constant_curvature,xiong2022pseudo} is essentially the {geodesic} curve on a hyperbolic manifold with an explicit formulation due to the manifold type restriction. We do not enforce any assumption here and let the model learn the embedding geometry.
\begin{algorithm}[!htb]
\caption{Graph Node Embedding with HamGNN}
\label{alg:hamgnn}
\begin{algorithmic}[1]
\STATE \textbf{Initialize:} the network modules including Hamiltonian function network $\{H^{(\ell)}_{\mathrm{net}}\}_{\ell=1}^{L}$, the learnable momentum function $\{ Q^{(\ell)}_{\mathrm{net}}\}_{\ell=1}^{L}$, and the raw node features compressor network FC.

\STATE \textbf{I. Training:}
\FOR{\text{Epoch} $1$ \textbf{to} $N$}
    \STATE \tb{1).} At each epoch, perform the following to obtain the embedding $Z$ with the $n$-th column being ${}^nq^{(L)}(T)$:
    \STATE \textbf{Input:} $\mathcal{G} = (\mathcal{V}, \mathcal{E})$ with raw node features. \\
    Apply FC to compress raw node features and get $\{{}^nq^{(1)}\}_{n=1}^{|\mathcal{V}|}$.
    \FOR{layer $\ell=1$ \textbf{to} $L$}
        \STATE \tb{i).} Solve the following Hamiltonian equations \cref{eq:ham_orb} (or its variants) using neural differential equation solvers \cite{chen2018neural}:\\
        \[
        \dot{q}^{(\ell)i}=\frac{\partial H^{(\ell)}_{\mathrm{net}}}{\partial p^{(\ell)}_i}, \quad \dot{p}^{(\ell)}_i=-\frac{\partial H^{(\ell)}_{\mathrm{net}}}{\partial q^{(\ell)i}}
        \]
        with $q^{(\ell)}(0)={}^nq^{(\ell)}$ and $p^{(\ell)}(0) = Q^{(\ell)}_{\mathrm{net}}({}^np^{(\ell)})$ at $t=0$. The feature vector ${}^nq^{(\ell)}(T)$ at time $T$ is obtained from this step.
        \STATE \tb{ii).} Perform neighborhood aggregation according to the following equation \cref{eq.agg} to obtain ${}^nq^{(\ell+1)}$:
        \[
        { }^n q^{(\ell+1)}={ }^n q^{(\ell)}(T)+\frac{1}{|\mathcal{N}(n)|} \sum_{m \in \mathcal{N}(n)}{ }^m q^{(\ell)}(T)
        \]
        which is used as the initial condition at the next layer $\ell+1$.
    \ENDFOR
    \STATE \tb{2).} Utilize backpropagation to minimize the cross-entropy loss in node classification and link prediction, with the latter employing negative sampling.
    \STATE \tb{3).} Perform validation over the validation split.
    \STATE \tb{4).} Save the model parameters.
\ENDFOR
\STATE \textbf{II. Testing:}
\STATE Load the model from the best validation epoch and perform \textbf{Step I.1).} to obtain the final embedding over the test split. Perform node classification or link prediction.
\end{algorithmic}
\end{algorithm}
\subsubsection{Learnable $H_{\mathrm{net}}$} \label{sec:H_1}
Different from \cref{sec:H_geo} where $H$ is set as \cref{eq:dae} with a pseudo-Riemannian metric, we choose a more flexible $H$ instantiated by an FC layer and consider the Hamiltonian equations: 
\begin{align}
\begin{aligned}
\dot{q}^i =\frac{\partial H_{\mathrm{FC}}}{\partial p_i}, \quad
\dot{p}_i =-\frac{\partial H_{\mathrm{FC}}}{\partial q^i}.\label{eq:H_1}
\end{aligned}
\end{align}
\subsubsection{Learnable Convex $H_{\mathrm{net}}$} \label{sec:H_2}
As discussed in \cref{ssec:pre_ham_orb}, if $H_{\mathrm{net}}$ is restricted to strictly convex functions, the Hamiltonian formalism  can degenerate to a \emph{Lagrangian formalism} through the Legendre transformation \cref{eq:Legendre}.
We take the following restricted Hamiltonian equations
\begin{align}
\begin{aligned}
\dot{q}^i =\frac{\partial H_{\mathrm{net}}}{\partial p_i},
\dot{p}_i  =-\frac{\partial H_{\mathrm{net}}}{\partial q^i}, 
\st &H_{\mathrm{net}} \text{ is convex},
\label{eq:H_2}
\end{aligned}
\end{align}
where a stationary action in \cref{eq:sq} is achieved. 
To guarantee that $H_{\mathrm{net}}$ is convex, we follow the work in \cite{amos2017input} to set non-negative layer weights from the second layer in $H_{\mathrm{net}}$, and all activation functions in $H_{\mathrm{net}}$ to be convex and non-decreasing. This network design is shown to be able to approximate any convex functions in \cite{chen2018optimal}.
\subsubsection{Learnable  $H_{\mathrm{net}}$ with Relaxation} \label{sec:H_4}
Different from \cref{sec:H_1}, we enforce additional system biases along the curve as follows:
\begin{align}
\begin{aligned}
\dot{q}^i =\frac{\partial H_{\mathrm{net}}}{\partial p_i}, \quad
\dot{p}_i =-\frac{\partial H_{\mathrm{net}}}{\partial q^i} + f_{\mathrm{net}}(q).\label{eq:H_4}
\end{aligned}
\end{align}
Instead of keeping the energy during the feature update along the Hamiltonian orbit, we now also include an additional energy term during the node feature update.
\subsubsection{Learnable $H_{\mathrm{net}}$ with a flexible symplectic form} \label{sec:H8}
Hamiltonian equations have a more flexible representation using the symplectic $2$-form (cf. \cref{appsec:app_diff}). The chart coordinate representation $(q,p)$ may not be the Darboux coordinate system for the symplectic $2$-form. Even if our learnable $H_{\mathrm{net}}$ may be able to learn the energy representation under the chosen chart coordinate system, we consider a learnable symplectic $2$-form to act in concert with $H_{\mathrm{net}}$.
More specifically, following \cite{ChenNeurIPS2021}, we have the following 1-form
\begin{align*}
    \theta^1_{\mathrm{net}} = f_{i,\mathrm{net}} \d q^i,
\end{align*}
where $ f_{i, \mathrm{net}}: M\to \R^d$ is the output's $i$-th component of the neural network parameterized function. The Hamiltonian orbit is then provided by \cite{ChenNeurIPS2021} as follows:
\begin{align}
   \left(\dot{q}^i, \dot{p}^i\right) = W^{-1}(q,p) \nabla H_{\mathrm{net}}(q,p), \label{eq:H_8}
\end{align}
where the skew-symmetric $2d\times 2d$ matrix $W$, whose elements are written in terms of $\left(\p_if_{j,\mathrm{net}} -\p_jf_{i,\mathrm{net}} \right)$, is given in \cref{appeq:W} in the appendix due to space limitations.

\section{Experiments}\label{sec:exp}

In this section, we implement the proposed HamGNNs with different settings as shown in \cref{sec:diff_ham} and \cref{appsec:app_formulations}.
We select \tb{datasets with various geometries} including the three citation networks: Cora \cite{McCallum2004AutomatingTC}, Citeseer \cite{SenNamata2008}, Pubmed \cite{namata:mlg12-wkshp}; and two low hyperbolicity datasets \cite{chami2019hyperbolic_GCNN}: Disease and Airport (cf. \cref{tab:bechmark}). 
Furthermore, we create \tb{new datasets with more complex geometry} by combining Disease and Cora/Citeseer so that the new datasets have a mixture of both hyperbolic and Euclidean local geometries.\\
Adhering to the experimental settings of \cite{chami2019hyperbolic_GCNN,gu2018learning,bachmann2020constant_curvature,lou2020differentiating,xiong2022pseudo}, we evaluate the effectiveness of the node embedding by performing \emph{two downstream tasks: node classification and link prediction} using the embeddings. Rather than beating all the existing GNNs on these two specific tasks, we want to demonstrate that our node embedding strategy is able to automatically learn, without extensive tuning, the underlying geometry of any given graph dataset even if its underlining geometry is very complex, e.g., a mixture of multiple geometries. Such examples can be found in \cref{tab:mix_geo}. Due to space constraints, we refer the readers to \cref{appsec:app_expset} for the description of datasets and implementation~details.

To fairly compare the performance of the proposed HamGNN, for the node classification tasks, we select several popular GNN models as the baseline.
These include \emph{Euclidean GNNs}: GCN \cite{kipf2016semi}, GAT \cite{velickovic2018graph}, SAGE \cite{hamilton2017inductive}, and SGC \cite{wu2019simplifying}; \emph{Hyperbolic GNNs} \cite{chami2019hyperbolic_GCNN, liu2019hyperbolic_GNN}: HGNN, HGCN, HGAT and LGCN \cite{zhang2021lorentzian}; a \emph{GNN that mixes Euclidean and hyperbolic embeddings}: GIL \cite{zhu2020GIL}; \emph{(Pseudo-)Riemannian GNNs}: $\kappa$-GCN \cite{bachmann2020constant_curvature} and 
$\calQ$-GCN \cite{xiong2022pseudo}; as well as \emph{Graph Neural Diffusions}: GRAND \cite{chamberlain2021grand} and GraphCON \cite{rusch2022icml}.
We also include the MLP baseline, which does not utilize the graph topology information.
To further demonstrate the advantage of HamGNN, we also include one vanilla ODE system, whose formulation is given in \cref{appsec:app_exp_van_ode}. This vanilla ODE system neither includes the learnable "momentum" vector $p$ nor adheres to the Hamiltonian orbits \cref{eq:H_3}.
For the link prediction task, we compare HamGNN to the \emph{standard graph node embedding models}, including all the aforementioned baselines in the node classification task except the graph neural diffusion baselines. 
For the link prediction tasks, we report the best results from different versions of HamGNN: HamGNN \cref{eq:H_geo} on the Disease dataset and \cref{eq:H_1} on the remaining datasets.

\begin{table}[ht]\small
\centering
\resizebox{0.5\textwidth}{!}{\setlength{\tabcolsep}{1pt} 
\begin{tabular}{c|c|c|c|c|ccc}
\toprule
Method  & Disease & Airport & Pubmed & Citeseer & Cora  \\
$\delta$-hyperbolicity  & {0.0} &{1.0} &{3.5} &{4.5} &{11.0}\\
\midrule
MLP  &  50.00$\pm$0.00  &  76.96$\pm$1.77  &  71.95$\pm$1.38  &  58.10$\pm$1.87  &  57.15$\pm$1.15 \\
HNN \cite{ganea2018hyperbolic_neural_net} &  56.40$\pm$6.32  &  80.49$\pm$1.54  &  71.60$\pm$0.47  &  55.13$\pm$2.04  &  58.03$\pm$0.55 \\
\midrule
GCN \cite{kipf2016semi} &  81.10$\pm$1.33  &  82.25$\pm$0.56  &  77.83$\pm$0.77  &  {71.78$\pm$0.34}  &  80.29$\pm$2.29 \\
GAT \cite{velickovic2018graph} &  87.01$\pm$2.77  &  92.99$\pm$0.83  &  77.58$\pm$0.81  &  68.07$\pm$1.31  &  80.33$\pm$0.61 \\
SAGE \cite{hamilton2017inductive} &  81.60$\pm$7.68  &  81.97$\pm$0.85  &  77.63$\pm$0.15  &  65.90$\pm$2.32  &  74.50$\pm$0.88 \\
SGC \cite{wu2019simplifying} &  82.78$\pm$0.93  &  81.40$\pm$2.21  &  76.83$\pm$1.11  &  70.88$\pm$1.32  &  81.98$\pm$1.71 \\
\midrule
HGNN \cite{liu2019hyperbolic_GNN} &  80.51$\pm$5.70  &  84.54$\pm$0.72  &  76.65$\pm$1.38  &  69.43$\pm$0.99  &  79.53$\pm$0.98 \\
HGCN \cite{chami2019hyperbolic_GCNN} &  89.87$\pm$1.13  &  85.35$\pm$0.65  &  76.38$\pm$0.81  &  65.80$\pm$2.04  &  78.70$\pm$0.96 \\
HGAT \cite{zhang2021hyperbolic_graph_att_net} &  88.68$\pm$3.36  &  87.50$\pm$0.99  &  78.00$\pm$0.50  &  69.20$\pm$0.96  &  80.88$\pm$0.75 \\
LGCN \cite{zhang2021lorentzian} &   88.47$\pm$1.80   &   88.22$\pm$0.18    &   77.35$\pm$1.38  &   68.08$\pm$1.98    &    80.60$\pm$0.92 \\
\midrule
GIL \cite{zhu2020GIL} &  90.78$\pm$0.45  &   91.52$\pm$1.74  &  77.76$\pm$0.57  &  71.10$\pm$1.24  &  82.10$\pm$1.12 \\
\midrule
$\kappa$-GCN \cite{bachmann2020constant_curvature}  & -    &87.92$\pm$1.33    & \third{79.20$\pm$0.65}  & \second{73.25$\pm$0.51} &81.08$\pm$1.45\\
$\calQ$-GCN \cite{xiong2022pseudo}  & 70.79$\pm$1.23    &89.72$\pm$0.52  & \first{81.34$\pm$1.54}  & \first{74.13$\pm$1.41} & \first{83.72$\pm$0.43 }\\
\midrule
Vanilla ODE in \cref{appeq:normal_ode}& 71.81$\pm$18.85    &  90.34$\pm$0.67  & 73.30$\pm$3.31   & 56.60$\pm$1.26   &   68.38$\pm$1.18   \\
\midrule
GRAND \cite{chamberlain2021grand} & 74.52$\pm$3.37 & 60.02$\pm$1.55 & \second{79.32$\pm$0.51} & 71.76$\pm$0.79 & \second{82.80$\pm$0.92}  \\
GraphCON \cite{rusch2022icml} & 87.50$\pm$4.06    & 68.61$\pm$2.10   & 78.80$\pm$0.97 & 71.33$\pm$0.83  &  82.49$\pm$1.08   \\

\midrule
\midrule
HamGNN \cref{eq:H_geo} &  \second{91.26$\pm$1.40}  &  \third{95.50$\pm$0.48} &  78.08$\pm$0.48  &  70.12$\pm$0.86  &  \third{82.16$\pm$0.80} \\
HamGNN \cref{eq:H_1} &  88.74$\pm$1.17  &  95.11$\pm$0.40  &  78.18$\pm$0.54  &  71.48$\pm$1.42  &  81.52$\pm$1.27 \\
HamGNN \cref{eq:H_2}  & 84.57$\pm$5.78  &  93.28$\pm$1.40  &  78.83$\pm$0.46 & 72.00$\pm$0.73  &  81.84$\pm$0.88 \\
HamGNN \cref{eq:H_4} &  87.48$\pm$5.90  &  95.46$\pm$0.68  &  78.30$\pm$0.34  &  \third{72.38$\pm$0.85}  &  81.56$\pm$0.97 \\
HamGNN \cref{eq:H_8} &  88.35$\pm$1.51  &  93.66$\pm$0.16  &  {78.60$\pm$0.32}  &  71.52$\pm$1.41  &  81.24$\pm$0.59 \\
HamGNN \cref{eq:H_6} &  \third{91.18$\pm$0.99}  &  \second{95.80$\pm$0.19}  &  77.90$\pm$0.49  &  69.18$\pm$1.63  &  80.90$\pm$0.35 \\
HamGNN \cref{eq:H_3} &  \first{91.50$\pm$2.07}  &  \first{95.99$\pm$0.13}  &  78.26$\pm$0.64  &  69.10$\pm$1.95  &  80.10$\pm$1.56 \\
\bottomrule
\end{tabular}}
\caption{Node classification accuracy(\%). The best, second best, and third best results for each criterion are highlighted in \first{red}, \second{blue}, and \third{cyan}, respectively. ``-'' indicates the open source code and/or the result is unavailable.}
\label{tab:noderesults}
\end{table}

\begin{table}[!htb]\small
\centering
\resizebox{0.5\textwidth}{!}{\setlength{\tabcolsep}{1pt} 
\begin{tabular}{c|c|c|c|c|ccc} 
\toprule
Method  & Disease & Airport & Pubmed & Citeseer & Cora  \\
$\delta$-hyperbolicity  & {0.0} &{1.0} &{3.5} &{4.5} &{11.0}\\
\midrule
MLP &  \third{83.37$\pm$5.04}  &  87.04$\pm$0.56  &  88.69$\pm$1.59  &  89.65$\pm$1.00  &  91.07$\pm$0.56 \\
HNN \cite{ganea2018hyperbolic_neural_net}&  81.37$\pm$8.78  &  86.06$\pm$2.08  &  94.69$\pm$0.25  &  89.83$\pm$0.39  &  92.83$\pm$0.76 \\
\midrule
GCN \cite{kipf2016semi} &  60.38$\pm$2.51  &  90.97$\pm$0.65  &  91.37$\pm$0.09  &  93.20$\pm$0.28  &  92.89$\pm$0.77 \\
GAT \cite{velickovic2018graph}  &  62.03$\pm$1.58              &  91.05$\pm$0.83  &  91.03$\pm$0.67        &  93.83$\pm$0.65  &  93.34$\pm$0.50 \\
SAGE \cite{hamilton2017inductive} &  68.02$\pm$0.43             &  91.40$\pm$0.88  &  93.61$\pm$0.26        &  93.37$\pm$0.88  &  92.94$\pm$0.40 \\
SGC \cite{wu2019simplifying} &  59.83$\pm$4.01                  &  89.72$\pm$0.82  &  92.16$\pm$0.13       &  94.78$\pm$0.77  &  93.15$\pm$0.22 \\
\midrule
HGNN \cite{liu2019hyperbolic_GNN} &  60.20$\pm$1.14           &  92.46$\pm$0.20  &  93.09$\pm$0.09            &90.35$\pm$0.57  &  92.05$\pm$0.33 \\
HGCN \cite{chami2019hyperbolic_GCNN} &  78.09$\pm$2.79        &  94.28$\pm$0.20   &  \third{96.79$\pm$0.01}  &93.60$\pm$0.14  &  94.10$\pm$0.05 \\
HGAT \cite{zhang2021hyperbolic_graph_att_net}&76.32$\pm$3.41    &  94.64$\pm$0.51  &  \first{96.86$\pm$0.03}  & 93.45$\pm$0.25  &  94.96$\pm$0.36 \\
\midrule
GIL \cite{zhu2020GIL}  &\first{99.97$\pm$0.08}                &\second{97.92$\pm$2.64}  & 91.22$\pm$3.25       &\third{95.99$\pm$8.89}  &  \second{97.78$\pm$2.31} \\
\midrule
$\kappa$-GCN\cite{bachmann2020constant_curvature} &-                                               &\third{96.35$\pm$0.62}  &96.60$\pm$0.32         &95.79$\pm$0.24  &94.04$\pm$0.34 \\
$\calQ$-GCN \cite{xiong2022pseudo} &-                                                &96.30$\pm$0.22          &\second{96.86$\pm$0.37}  &\second{97.01$\pm$0.30} &\third{95.16$\pm$1.25}\\
\midrule
HamGNN  & \second{99.73$\pm$0.26}                          &\first{99.99$\pm$0.01}     & 92.15$\pm$0.30        & \first{99.99$\pm$0.00}     &\first{98.20$\pm$1.73}    \\
\bottomrule
\end{tabular}}
\caption{Link prediction ROC(\%). The best, second best, and third best results for each criterion are highlighted in \first{red}, \second{blue}, and \third{cyan}, respectively. ``-'' indicates the open source code and/or the result is not available.}
\label{tab:linkresults}
\end{table}

\subsection{Performance Results and Ablation Studies}\label{sec:per}
\tb{Node classification.} The  node classification performance on the benchmark datasets  using the baseline models and the proposed HamGNNs with different $H_{\mathrm{net}}$s is shown in \cref{tab:noderesults}. 
We observe that HamGNN adapts well to all datasets with various geometries. 
For the datasets such as Cora, Citeseer and Pubmed, which can be well embedded into Euclidean space, HamGNNs achieve comparable performance to the conventional Euclidean GNNs. 
For example, HamGNN \cref{eq:H_geo} achieves the third-best performance on the Cora. 
For tree-like graph data Disease and Airport, HamGNNs outperform the other GNNs, including all the hyperbolic GNNs, which are tailored for hyperbolic datasets, and Riemannian GNNs which use a Cartesian product of spherical, hyperbolic, and Euclidean spaces. 
From the results in \cref{tab:noderesults}, we can see that Riemannian GNNs $\kappa$-GCN and $\calQ$-GCN are biased towards embedding data into Euclidean space since they have the best performance on high hyperbolicity datasets but average performance on low hyperbolicity datasets.\\
Furthermore, the comparison between HamGNNs with the vanilla ODE GNN in \cref{appeq:normal_ode} without any Hamiltonian mechanism implies the importance of the Hamiltonian layer.\\
\tb{Link prediction.} In \cref{tab:linkresults}, we report the averaged ROC for the link prediction task. We observe HamGNN adapts well to all datasets and is the best performer on the Airport, Citeseer, and Cora. \\
\tb{Comparison between different $H_{\mathrm{net}}$.} We compare HamGNNs with different $H_{\mathrm{net}}$s as elaborated in \cref{sec:diff_ham} on the node classification task. In \cref{ssec:pre_ham_orb}, we argue that the geodesic curves derived from the action of curve length  may potentially sacrifice efficacy for the graph node embedding task since we do not know what a reasonable action formulation that guides the evolution of the node feature in this task is. 
We therefore also include a more flexible $H_{\mathrm{net}}$ in \cref{eq:H_1} with other variations, e.g., \cref{eq:H_2,eq:H_4,eq:H_8}. 
From \cref{tab:noderesults}, we observe that the original HamGNN \cref{eq:H_geo} shows good adaptations of node embedding to various datasets. 
Simply using a FC layer for $H$ in \cref{eq:H_1} has no obvious improvement. Further imposing convexity on $H$ in \cref{eq:H_2} also has little positive impact on the performance.  
Including system bias in \cref{eq:H_4} achieves the best performance on Citeseer among all HamGNNs. 
The HamGNNs that achieve the best performance on Disease and Airport are \cref{eq:H_6} and \cref{eq:H_3}. 
Overall, all HamGNN variants elaborated in  \cref{sec:diff_ham} have well-adapted node embedding performance for various datasets  even though some variants may perform slightly better. This observation indicates that good geometry adaptations may come from the HamGNN model architecture with the Hamiltonian orbit evolution. This is further verified by the node classification performance from vanilla ODE in \cref{appeq:normal_ode}, which does not have a well-adapted node embedding performance and is designed without the philosophy of Hamiltonian mechanics. Good experiment results in \cref{sec.mixx} using both \cref{eq:H_geo,eq:H_1} also provide further evidence to support our conclusion.\\
We next compare the HamGNNs using \cref{eq:H_1} and \cref{eq:H_8}. The difference between those two settings is that  the symplectic form in \cref{eq:H_1} is set to be the special Poincar\'e $2$-form while in \cref{eq:H_8}, the symplectic form is a learnable one. We however observe that the two HamGNNs achieve similar performance and the more flexible symplectic form does not improve the model performance. This may be because of the fundamental Darboux theorem \cite{lee2013smooth} in symplectic geometry, which states that we can always find a Darboux coordinate system to give any symplectic form the \emph{Poincar\'e $2$-form}. The feature compressing FC may have the network capacity to approximate the Darboux coordinate map, while the flexible learnable $H_{\mathrm{net}}$ also has the network capacity to get the energy representation under the chosen chart coordinate system.

\begin{table}[!htb]\small
    \centering
 \resizebox{0.5\textwidth}{!}{\setlength{\tabcolsep}{1pt}
 \begin{tabular}{c|c|c|c|c|c|cc}
    \hline
        datasets & GCN & HGCN & GIL & $\calQ$-GCN  & HamGNN \cref{eq:H_geo} & HamGNN \cref{eq:H_1}\\
        \hline
        Air+Cora & 75.16$\pm$0.65 & 78.72$\pm$0.42  & 82.04$\pm$1.27  & 90.25$\pm$0.81 & \tb{94.82$\pm$0.78} & 93.08$\pm$0.55\\
          \hline
               Air+Cite & 70.97$\pm$0.34 & 74.65$\pm$0.40  &  77.83$\pm$1.57  & 87.63$\pm$0.28 & \tb{90.03$\pm$0.29} &  88.94$\pm$0.75   \\
         \hline
    \end{tabular}}
    \caption{Node classification on the Mixed Geometry Dataset. First row: mixed Airport+Cora dataset; Second row: mixed Airport+Citeseer dataset.}
    \label{tab:mix_geo}
\end{table}

\begin{table}[!htp]\small
\centering
\resizebox{0.5\textwidth}{!}{
\begin{tabular}{lcccccccc}
\toprule
Dataset  &  Models  &  3 layers   &  5 layers   &  10 layers   &  20 layers  \\
\midrule
\multirow{3}{*}{Cora} 
& GCN &  80.29$\pm$2.29  &  69.87$\pm$1.12  &  26.50$\pm$4.68  &   23.97$\pm$5.42  \\
& HGCN &  78.70$\pm$0.96  &  38.13$\pm$6.20  &  31.90$\pm$0.00  &    26.23$\pm$9.87  \\
& HamGNN \cref{eq:H_2} &  \tb{81.84$\pm$0.88}  &  \tb{81.08$\pm$0.16}  &  \tb{81.40$\pm$0.44}  &   \tb{80.58$\pm$0.30} \\
\bottomrule
\end{tabular}}
\caption{Node classification accuracy(\%) when increasing the number of layers on the Cora dataset.}
\label{tab:num_layer_cora}
\end{table}
\subsection{Mixed Geometry Dataset}\label{sec.mixx}
\subsubsection{Mixed Airport + Cora Dataset} 
To understand the node embedding capacity of HamGNN, we created a new mixed geometry graph dataset by combining the Airport (3188 nodes) and Cora (2708 nodes) datasets into a single large graph with 5896 nodes (cf.\ \cref{tab:bechmark}). The results are presented in the first row of \cref{tab:mix_geo}.\\
This new dataset is composed of a mixture of hyperbolic and Euclidean geometries, as the Airport dataset is known to have a more hyperbolic geometry, and the Cora dataset is known to have a more Euclidean geometry, as shown in \cref{fig:hyper}. We choose these two datasets also because they have a similar number of nodes. 
To standardize the node feature dimension across datasets, we pad the node features with additional zeros.
We do not create any \emph{new} edges in the new dataset so that the new graph contains two disconnected subgraphs: Airport and Cora. 
We use a 60\%, 20\%, 20\% random split for training, validation, and test sets on this new dataset.\\
We test our HamGNN  against several baselines including GCN, HGCN, GIL, and $\calQ$-GCN. The results are shown in \cref{tab:mix_geo} from which we can observe that, in this more complex geometry setting, HamGNN performs the best. 
\subsubsection{Mixed Airport + Citeseer Dataset} 
We also include experiments on another new mixed dataset created using  Airport and Citeseer. The combination of these two datasets is the same as the previous one. 
This new dataset therefore has different geometry in each component from Airport or Citeseer.
The results are reported in the second row of \cref{tab:mix_geo}. We observe that our HamGNN still performs the best. 

These two experiments show that our model can adapt well to the underlying complex geometry.
\subsection{Observation of Resilience to Over-Smoothing}\label{sec:ove_smo}
As a side benefit of HamGNN, we observe from \cref{tab:num_layer_cora} that if more Hamiltonian layers are stacked, HamGNN is still able to distinguish nodes from difference classes, while the other GNNs suffer severe oversmoothing problem \cite{chen2020measuring}. 
This may be because when updating node features their energy is constrained to be alone in the Hamiltonian orbit. So, if two nodes are distinct in terms of their energy at the input of HamGNN, they will still be distinguishable by their energy at the output of HamGNN, no matter how many times the feature aggregation operation has been implemented.

\tb{More Experiments.} We kindly refer readers to \cref{appsec:app_expmore} where we include more empirical analysis and visualization.
\section{Conclusion}
In this paper, we have designed a new node embedding strategy from Hamiltonian orbits that can automatically learn, without extensive tuning, the underlying geometry of any given graph dataset even when multiple different geometries coexist. 
We have demonstrated empirically that our approach adapts better than popular state-of-the-art graph node embedding GNNs to various graph datasets on two graph node embedding downstream tasks.


\section*{Acknowledgments}
This research is supported by A*STAR under its RIE2020 Advanced Manufacturing and Engineering (AME) Industry Alignment Fund – Pre Positioning (IAF-PP) (Grant No. A19D6a0053) and the National Research Foundation, Singapore and Infocomm Media Development Authority under its Future Communications Research and Development Programme.
The computational work for this article was partially performed on resources of the National Supercomputing Centre, Singapore (https://www.nscc.sg).


\clearpage
\bibliography{IEEEabrv,icml2023bib}
\bibliographystyle{icml2023}
\newpage
\appendix

\section{Related Work}\label{appsec:related}
Graph Neural Networks (GNNs) \cite{yueBio2019, AshoorNC2020, kipf2017semi, ZhangTKDE2022, WuTNNLS2021, LeeJiTay:C22, ruishe23, WanKanShe:C23, JiLeeMen:C23} have exhibited remarkable inferential performance in a broad range of applications. In this section, we provide a succinct review of Hamiltonian Neural Networks, Riemannian Manifold GNNs, and Graph Neural Diffusions.

\textbf{Hamiltonian neural networks.}
Among the physics-inspired deep learning approaches, Hamiltonian equations have been applied to conserve an energy-like quantity when training neural networks. The work by \cite{haber2017stable} introduces a Hamiltonian-inspired neural ODE to stabilize gradients, avoiding vanishing and exploding issues. Further,  \cite{huang22a} investigates the adversarial robustness of Hamiltonian ODE. In the physics community, several works propose learning a Hamiltonian function from the observation of systems to simulate physical systems or solve problems in particle physics. Studies such as \cite{greydanus2019hamiltonian, zhongiclr2020, ChenNeurIPS2021} train a neural network to infer the Hamiltonian dynamics of a physical system, where the Hamiltonian equations are solved using neural ODE solvers. To better model interactions between physical objects, works like \cite{bishnoi2022enhancing,sanchezhamiltonian} employ a graph network with a Hamiltonian inductive design to capture the dynamics of physical systems from observed trajectories. For instance, to forecast dynamics, \cite{bishnoi2022enhancing} employs graph networks incorporating Hamiltonian mechanics to learn phase space orbits efficiently. This work demonstrates the effectiveness of Hamiltonian graph neural networks on several dynamics benchmarks, including pendulum systems and spring networks. 
Compared to traditional neural networks, Hamiltonian (graph) neural networks offer several advantages, such as energy preservation, natural incorporation of conservation laws, and a more interpretable and physically meaningful representation of the system.

\emph{In this paper, we are neither simulating a Hamiltonian physics system nor solving a real-world Hamiltonian physics problem.
Instead, we apply Hamiltonian equations to generic graph node embedding tasks. Our approach generalizes the Riemannian manifold GNNs and has not been investigated in the above-mentioned works.}

\tb{Riemannian manifold GNNs.}
Most GNNs in the literature \cite{yueBio2019,AshoorNC2020,kipf2017semi,ZhangTKDE2022,WuTNNLS2021} embed graph nodes in Euclidean spaces. In what follows, we simply call them (vanilla or Euclidean) GNNs. They perform well on some datasets like the Cora dataset \cite{McCallum2004AutomatingTC} whose Gromov $\delta$-hyperbolicity distributions \cite{chami2019hyperbolic_GCNN} is high. When dealing with datasets whose $\delta$-hyperbolicities are low (hence embedding should more appropriately be in a hyperbolic space) such as the Disease \cite{chami2019hyperbolic_GCNN} and Airport \cite{chami2019hyperbolic_GCNN} datasets, those GNNs suffer from improper node embedding.
To better handle hierarchical graph data, \cite{liu2019hyperbolic_GNN,chami2019hyperbolic_GCNN,zhang2021lorentzian,zhu2020GIL,WanKanSheWan:C23} propose to embed nodes into a hyperbolic space, thus yielding hyperbolic GNNs. Moreover, \cite{zhu2020GIL} proposes a mixture of embeddings from Euclidean and hyperbolic spaces. This mixing operation relaxes the strong space assumption of using only one type of space for a dataset. 
In some recent studies, such as \cite{gu2018learning,bachmann2020constant_curvature,lou2020differentiating}, the researchers use (products of) \emph{constant curvature} Riemannian spaces for graph node embedding where the spaces are assumed to be spherical, hyperbolic, or Euclidean. In work by \cite{xiong2022pseudo}, a special type of pseudo-Riemannian manifold called the pseudo-hyperboloid, which has constant non-zero curvature and is diffeomorphic to the product of a unit sphere and Euclidean space, has been considered for the same purpose.

\emph{In this paper, we embed nodes into a general learnable manifold via the Hamiltonian orbit on its symplectic cotangent bundle.} This allows our model to flexibly adapt to the inherent geometry of the dataset.

\tb{Graph neural diffusion.}
Neural Partial Differential Equations (PDEs) \cite{chamberlain2021grand,chamberlain2021blend,SonKanWan:C22,ZhaKanSon:C23} are extensions of neural Ordinary Differential Equations (ODEs) \cite{chen2018neural,kang2021Neurips} and have been applied to graph-structured data, where different diffusion schemes are assumed when performing message passing on graphs. To be more specific, the heat diffusion model is assumed in \cite{chamberlain2021grand} and the Beltrami diffusion model is assumed in \cite{chamberlain2021blend,SonKanWan:C22}. \cite{rusch2022icml} models the nodes in the graph as coupled oscillators, i.e., a second-order ODE. 

\emph{While the above-mentioned graph neural diffusion schemes and our model all use ODEs, there is a fundamental difference between our model and graph neural flows. The graph PDE models wrap the message passing function, e.g., aggregation functions like the one in GCN, and attention-based aggregation functions like the one in GAT, into an ODE function. In contrast, our model treats the node embedding process and node aggregation process as two independent processes: we use the ODE function only to learn a suitable node embedding space which is then followed by a node aggregation step. In summary, the ODE layer in our model is a node embedding layer taking node features as the input, whereas graph PDE layers can be interpreted as node aggregation layers taking node features as well as the graph adjacency matrix as the input.}\\

\section{Motivation} \label{sec.mot}
Our primary goal is to develop a more flexible and robust method for graph node embedding by leveraging the concepts of geodesic curves and Hamiltonian orbits on manifolds.

The embedding strategies in our work and the literature \cite{chami2019hyperbolic_GCNN,bachmann2020constant_curvature,xiong2022pseudo} can be unified in equation \cref{eq:sq} (and its dual equation \cref{eq:Legendre}):
$$S(q)=\int_a^b L(q(t), \dot{q}(t)) \mathrm{d} t.$$
Our work and the previous works all follow the same ``principle of stationary action'' (action means ``cost'', see below for more explanation) to minimize the functional \cref{eq:sq}. The minimization leads to different curves on fixed or arbitrary manifolds. The embedding strategy is to learn to move the node on the manifolds to specific positions by following these curves.

In previous works \cite{chami2019hyperbolic_GCNN,bachmann2020constant_curvature,xiong2022pseudo}, $L$ takes the \cref{eq:Lgeo}:
$$L=\frac{1}{2}\|\dot{q}(t)\|_{g(q(t))}^2=\frac{1}{2} g_{i k}(q) \dot{q}^i \dot{q}^k$$
where $g$ is fixed to some formulation, depending on whether the fixed manifold is assumed to be spherical, hyperbolic, or Euclidean. These methods learn to move the node on a fixed manifold following limited geodesic curves induced from a fixed $g$.
With the Lagrangian $L$ setting as \cref{eq:Lgeo}, we obtain the shortest geodesic that induces the exponential map in (pseudo-)Riemannian manifolds, which {has been widely used in the literature} \cite{chami2019hyperbolic_GCNN,bachmann2020constant_curvature,xiong2022pseudo} to map features to specific manifolds. These successes in the literature {demonstrate that the shortest geodesic or more general the ``principle of stationary action'' helps the node embedding quality.}

One philosophical reason for the success, as opined by Pierre Louis Maupertuis, is that ``nature is thrifty in all its actions'' \url{https://en.wikipedia.org/wiki/Stationary-action_principle} \cite{kline1990mathematical}. Here, actions mean ``effort'' or ``cost''. The classical physics example is that ``light travels between two given points along the path of shortest time''.
The embedding process (the feature evolution following a geodesic curve) on a manifold also obeys a similar least-action philosophy.

\textbf{Motivation I: Enhancing the adaptability of node embeddings using Riemannian geodesics.} Traditional graph node embedding approaches \cite{chami2019hyperbolic_GCNN,bachmann2020constant_curvature,xiong2022pseudo} often assume a fixed geometry $g$ that induces fixed Riemannian geodesics (i.e., exponential map) on manifolds for node embedding, which may not adequately represent all types of geometries in graph datasets. By integrating learnable geodesic curves with learnable $g$ on arbitrary (pseudo-)Riemannian manifolds, our method can adaptively learn the local geometry for each dataset. This not only enables the use of a broader class of functions but also offers improved representation for diverse graph data with varying underlying geometries.

In summary, for Motivation I, we replace the fixed $g$ in the aforementioned $L$ with a more flexible and learnable function. This learnable $g$ guides more adaptable geodesic curves on arbitrary (pseudo-)Riemannian manifolds, allowing us to learn the local geometry for each dataset more effectively. As a result, the graph node features that move along these learnable geodesic curves to positions on the manifold will lead to improved embedding quality.

\textbf{Motivation II: Enhancing node feature evolution with Hamiltonian orbits.} While the geodesic curve-based node embedding presented in Motivation I, using \cref{eq:Lgeo} with learnable $g$, allows for a more flexible representation of data, it still has limitations w.r.t.\ the generality of the node feature evolution along the manifold curve. To tackle this issue, we propose to extend the learnable geodesic curve concept to learnable Hamiltonian orbits on manifolds. These orbits, which are also curves, are associated with a more general Lagrangian function $L$. Our method can therefore effectively capture the underlying complexities and variations in the data, resulting in better performance in various graph learning tasks.

In summary, Motivation II involves further relaxing the aforementioned $L$ \cref{eq:Lgeo} to include more general functions, leading to more general Hamiltonian orbits (which are also curves) on manifolds. Note that geodesic curves are a specific type of Hamiltonian orbit (after the projection). As $L$ and $H$ are dual, we set different learnable $H$ in \cref{sec:diff_ham} of the paper. Graph node features continue to move along the learned curves to positions on the manifold, but the generalized $L$ allows for more flexible curve options than those provided in Motivation I.

\section{Some Formulations and More Hamiltonian Orbits}\label{appsec:app_formulations}
In section, we first present the formulations which are not shown in detail in the main paper due to space constraints. More Hamiltonian-related flows are also presented, which however do not strictly follow the Hamiltonian orbits on the cotangent bundle $T^*M$.

\subsection{$W$ in \cref{sec:H8}}
See \cref{appeq:W}.
\begin{figure*}[!tb]
\begin{equation}
\scalebox{1}{
   $W=\left(\begin{array}{cccc}0 & \p_1f_{2,\mathrm{net}}-\p_2f_{1,\mathrm{net}} & \p_1f_{3,\mathrm{net}}-\p_3f_{1,\mathrm{net}} & \cdots \\ \p_2f_{1,\mathrm{net}}-\p_1f_{2,\mathrm{net}} & 0 & \p_2f_{3,\mathrm{net}}-\p_3f_{2,\mathrm{net}} & \cdots \\ \p_3f_{1,\mathrm{net}}-\p_1f_{3,\mathrm{net}}& \p_3f_{2,\mathrm{net}}-\p_2f_{3,\mathrm{net}} & 0 & \cdots \\ \vdots & \vdots & \vdots & \ddots\end{array}\right) $\label{appeq:W}
   }
\end{equation}
\hrulefill
\end{figure*}

\subsection{Learnable Metric $g_{\mathrm{net}}$ with Relaxation} \label{sec:H_6}
Similar to \cref{sec:H_4}, we now impose additional system biases along the curve compared to the cogeodesic orbits \cref{sec:H_geo},
  \begin{align}
\begin{aligned}
\dot{q}^i =g_{\mathrm{net}}^{i j} p_j, \quad
\dot{p}_i = -\frac{1}{2} \p_i g_{\mathrm{net}}^{j k} p_j p_k+  f_{\mathrm{net}}(q).\label{eq:H_6}
\end{aligned}
\end{align}
Therefore, the projection of the curve from \cref{eq:H_6}  now  no longer follows  the geodesic curve along the base manifold equipped with metric $g_{\mathrm{net}}$.

\subsection{Hamiltonian Relaxation Flow with Higher Dimensional ``Momentum''}
In the paper main context, we present a new type of Hamiltonian-related flow, which does not strictly follow the Hamiltonian equations. Inspired from the work \cite{haber2017stable}, we now associate to each node $q\in \R^d$ an additional a {learnable} momentum vector $p\in \R^k$ which however is not strictly a cotangent vector of the manifold if $d\ne k$. We update the node features using the following equations
\begin{align}
    \begin{aligned}
        \dot{q} & = \phi \left(h^1_{\mathrm{net}}(p) - \rho q\right), \\
\dot{p} & =\phi \left(h^2_{\mathrm{net}}(q) - \rho p\right).\label{eq:H_3}
    \end{aligned}
\end{align}
where $h^1_{\mathrm{net}}$ and $h^2_{\mathrm{net}}$ are neural networks with $d$-dimensional output and $k$-dimensional output respectively, $\phi$ is a non-linear activation function and $\rho$ is a scalar hyper-parameter.

\section{Dataset Configuration}
The statistics of the datasets we use are reported in \cref{tab:bechmark}. We follow the data pre-processing strategy in \cite{chami2019hyperbolic_GCNN} to normalize the adjacency matrix and features before inputting them into the GNN models.

\begin{table}[ht]
\centering
\begin{tabular}{lcccc}
\toprule
 Dataset & Nodes & Edges & Classes & Node Features   \\
 \midrule
 Disease &   1044    &   1043    &    2      &   1000     \\

 Airport &   3188    &  18631     &     4     &     4   \\
 
 Cora &     2708  &    5429   &     7     &    1433    \\

 Citeseer &    3327   &   4732    &    6      &   3703     \\

 Pubmed &  19717     &  44338     &      3   &  500      \\
 \bottomrule
 
\end{tabular}
\caption{Dataset statistics, where the dataset hyperbolicity is shown in \cref{fig:hyper}.}
\label{tab:bechmark}
\end{table}

\section{Main Paper Experiments Setting} \label{appsec:app_expset}
We select the citation networks Cora \cite{McCallum2004AutomatingTC}, Citeseer \cite{SenNamata2008}, and Pubmed \cite{namata:mlg12-wkshp}, and the low-hyperbolicity \cite{chami2019hyperbolic_GCNN} Disease, Airport as the benchmark datasets. 
The citation datasets are widely used in graph representation learning tasks. We use the same dataset splitting settings in \cite{kipf2016semi}. The low-hyperbolicity datasets Disease and Airport are proposed in \cite{chami2019hyperbolic_GCNN}, where the Euclidean GNN models cannot learn the node embeddings effectively. We adhere to the data splitting and pre-processing procedures outlined in \cite{chami2019hyperbolic_GCNN} for the Disease and Airport datasets. 

We adjust the model parameters in HamGNN based on the results from the validation data. We use the ADAM optimizer \cite{kingma2014adam} with the weight decay as 0.001. We set the learning rate as 0.01 for citation networks and 0.001 for Disease and Airport datasets. 
 The results presented in \cref{tab:noderesults} are under the 3 layers HamGNN setting. We report the results by running the experiments over 10 times with different initial random seeds. 
 
 HamGNN first compresses the dimension of input features to the fixed hidden dimension (e.g. 64) through a fully connected (FC) layer. Then the obtained hidden features are input to the stacked  $H_{\mathrm{net}}$ ODE layers and aggregation layers. The $q$ in Hamiltonian flow is initialized by the node embeddings after the FC layer. 
\cref{tab:hlayers} shows the implementation details of the layers in HamGNN.

\subsection{ODE solver for Hamiltonian equations}
We employ the ODE solver \cite{torchdiffeq} in the implementation of HamGNN. 
For computation efficiency and performance effectiveness, the fixed-step explicit Euler solver \cite{chen2018neural} is used in HamGNN. We also compare the influence of ODE solvers and report the results in \cref{tab:odesolver}.
One drawback of the ODE solvers provided in \cite{torchdiffeq} is that they are not guaranteed to have the energy-preserving property in solving the Hamiltonian equations. However, this flaw does not significantly deteriorate our model performance regarding the embedding adaptation to datasets with various structures. Our extensive experiments on the node classification and link prediction tasks have demonstrated that the solvers provided in \cite{torchdiffeq} are sufficient for our use.
We leave the use of Hamiltonian equation solvers for future work to investigate whether solvers with the energy-preserving property can better help graph node embedding or mitigate the over-smoothing problem.

\begin{table*}[ht]
    \centering
    \begin{tabular}{ccccccc}
    \toprule
         ODE solver & Euler& Euler & Implicit Adams &  Implicit Adams & Dopri5  \\
    \midrule
    
         Step size & 0.1 & 0.5  & 0.1 &  0.5 & -  \\
        \midrule
         Cora & 81.10$\pm$1.13  & 81.52$\pm$1.27  & 81.62$\pm$0.58 &  81.40$\pm$0.77 & 81.62$\pm$0.58  \\
    \bottomrule

    \end{tabular}
    \caption{Node classification accuracy(\%) under different ODE solvers in HamGNN \cref{eq:H_1}. }
    \label{tab:odesolver}
\end{table*}

\section{More Ablation Studies and Experiments} \label{appsec:app_expmore}

\subsection{Vanilla ODE}\label{appsec:app_exp_van_ode}

To demonstrate the advantage of HamGNN's design, we also conduct more experiments that replace the Hamiltonian layer in HamGNN with a vanilla ODE as follows:
\begin{align}
    \dot{q}(t) = \tilde{f}_{\mathrm{net}}(q(t)) \label{appeq:normal_ode}
\end{align}
where the $\tilde{f}_{\mathrm{net}}(q(t)) $ is composed of two FC layers and a non-linear activation function. Compared to the Hamiltonian orbits in \cref{sec:diff_ham}, the equation \cref{appeq:normal_ode} does not include the learnable  ``momentum'' vector for each node and does not follow the   Hamiltonian orbits on the cotangent bundle.

\subsection{Influence of Metric Signature}
We vary the parameter $s$ to explore the influence of the signature of the learned metric $g$.  The results are presented in \cref{tab:signature}. We observe that for the Airport dataset, the influence of the signature is moderate where the best (with $r=4$) and worst performance (with $r=2$) has a moderate gap of 0.39\% in accuracy. For other datasets, metrics with different pre-defined signatures perform similarly to each other. In \cref{tab:noderesults} of the paper, we present the best signature. 
Further theoretical understanding of the influence of the signature is left for future work.

\begin{table}[!htb]
    \centering
    \setlength{\tabcolsep}{1pt}
 \resizebox{0.5\textwidth}{!}{\begin{tabular}{c|c|c|c|c|cc}
    \hline
        metric signature $(r,s)$ & Disease &  Airport & Pubmed & Citeseer & Cora\\
        \hline
         $(0,64)$ &  90.16$\pm$1.15 & 95.15$\pm$0.53 & 77.82$\pm$0.21  & 69.52$\pm$0.51  &  81.62$\pm$1.61   \\
         $(1,63)$ &  89.45$\pm$1.53  & 95.42$\pm$0.54 & 77.38$\pm$0.34  & 69.98$\pm$1.17  &  \textbf{82.16$\pm$0.80}   \\
         $(2,62)$ & 89.84$\pm$1.38   & 95.11$\pm$0.40 & 77.50$\pm$0.39  & 69.26$\pm$0.87  &  81.22$\pm$1.56   \\
         $(4,60)$ & 91.18$\pm$1.32   & \textbf{95.50$\pm$0.48} & 77.40$\pm$0.51  & 69.12$\pm$1.71  &  81.24$\pm$1.97   \\
         $(8,56)$ &  89.21$\pm$3.09  & 95.19$\pm$0.31 & 77.34$\pm$0.39  & 69.68$\pm$0.10  &  80.98$\pm$1.41   \\
         $(16,48)$ & 88.19$\pm$4.92  & 95.19$\pm$0.62 & \textbf{78.08$\pm$0.48}  & 69.52$\pm$0.97  & 80.86$\pm$1.26    \\
         $(32,32)$ & \textbf{91.26$\pm$1.40}  & 95.27$\pm$0.52 & 77.82$\pm$0.22  & \textbf{70.12$\pm$0.86}  &  81.32$\pm$1.06   \\
         \hline
    \end{tabular}}
    \caption{The impact of the signature of metric $g$ in on the node classification performance.}
    \label{tab:signature}
\end{table}

\subsection{Node Classification on Heterophilic Datasets}

\begin{table}[!htp]
    \centering
    \resizebox{0.4\textwidth}{!}{
    \begin{tabular}{ccccccccc}
    \hline
        Method & Cornell & Wisconsin & Texas  \\ \hline
        Geom-GCN & 60.54$\pm$3.67 & 64.51$\pm$3.66 & 66.76$\pm$2.72  \\ \hline
        H2GCN & 82.70$\pm$5.28 & 87.65$\pm$4.98 & 84.86$\pm$7.23 \\ \hline
        GPRGCN & 78.11$\pm$6.55 & 82.55$\pm$6.23 & 81.35$\pm$5.32 \\ \hline
        FAGCN & 76.76$\pm$5.87 & 79.61$\pm$1.58 & 76.49$\pm$2.87 \\ \hline
        GCNII & 77.86$\pm$3.79 & 80.39$\pm$3.40 & 77.57$\pm$3.83 \\ \hline
        MixHop & 73.51$\pm$6.34 & 75.88$\pm$4.90 & 77.84$\pm$7.73 \\ \hline
        WRGAT & 81.62$\pm$3.90 & 86.98$\pm$3.78 & 83.62$\pm$5.50 \\ \hline
        GraphCON & 75.14$\pm$4.95 & 84.90$\pm$2.64 & 80.00$\pm$3.66 \\ \hline
        HamGNN & 76.49$\pm$5.10 & 83.92$\pm$4.87 & 81.62$\pm$6.22 \\ \hline
    \end{tabular}}\label{tab:noderesultsheterophilic}
    \caption{Node classification accuracy(\%) on heterophilic datasets under 10 fixed 48\%/32\%/20\% splits taken from \cite{pei2020geom}.}
\end{table}

In this section, we include experiments on the node classification task using heterophilic graph datasets. 
We would like to emphasize that our HamGNN is not specifically designed for heterophilic datasets, where nodes with different attributes or classes are more likely to be linked together in the graph. Our primary focus is on homophily datasets with varying local geometry, and the experiments presented in the paper mainly address this focus. It is important to note that the local uniform message-passing aggregation \cref{eq.agg} used in our paper is also more suited to homophily datasets. The additional experiments on heterophilic datasets in the appendix serve to demonstrate that even in such cases, our model can achieve competitive performance when compared to GNNs that are specifically designed for heterophilic datasets, provided that a good node embedding layer is designed in our paper. In the context of heterophilic datasets, common techniques like higher-order neighbor mixing, adaptive message aggregation, and ego-neighbor separation \cite{zhu2020beyond,bo2021beyond,abu2019mixhop} hold the potential to enhance our model's performance. We intend to investigate these strategies and their impact on our novel node embedding approach in future work.

In this section, we select the heterophilic graph datasets Cornell, Texas and Wisconsin from the CMU WebKB \footnote{\url{http://www.cs.cmu.edu/afs/cs.cmu.edu/project/theo-11/www/wwkb/}} project where randomly generated splits of data are provided by \cite{pei2020geom}. The edges in these graphs represent the hyperlinks between webpages nodes. The labels are manually selected into  five classes, student, project, course, staff, and faculty. The features on node are the bag-of-words of the web pages. 

For the heterophilic graph datasets, we include the baselines GCN, GAT, SAGE, APPNP \cite{KlicperaBG19}, GCNII \cite{chen2020simple}, GPRGNN \cite{chien2020adaptive}, and H2GCN \cite{zhu2020beyond} which are the common baselines for heterophilic graph datasets \cite{bi2022make}. Additionally, we also include GraphCON \cite{rusch2022icml} for comparisons. We report the results by running the experiments over 10 times with different initial random seeds for GraphCON and HamGNN, while for the other baselines on the heterophilic graph datasets, we use the results reported in the paper \cite{luan2022revisiting}. We still observe that our method is competitive for \emph{GNNs that are designed specifically for heterophilic datasets.}

\subsection{Aggregation}

In the main paper, we have chosen to implement a simple yet effective fixed-weight aggregation in accordance with Occam's razor principle rather than incorporating complex and computationally heavy attention-based aggregation operations. However, our model is compatible with more advanced aggregation methods. To demonstrate this, we have conducted additional experiments using attention-based aggregation with \cref{eq:H_geo}, and the results are presented in \cref{tab:aggregationatt}. We observe that on the Disease dataset, the attention-based aggregation can further improve the node classification accuracy. 

\begin{table}[!htb]
    \centering
    \setlength{\tabcolsep}{1pt}
 \resizebox{0.5\textwidth}{!}{\begin{tabular}{c|c|c|c|c|cc}
    \toprule
        Aggregation & Disease & Airport & Pubmed & Citeseer & Cora\\
        \hline
         attention &  92.72$\pm$1.65 & 94.35$\pm$1.00 & 78.03$\pm$0.69  & 70.53$\pm$1.84  &  81.91$\pm$1.03   \\
         \hline
          fixed-weight \cref{eq.agg}  & 91.26$\pm$1.40  &  95.50$\pm$0.48 &  78.08$\pm$0.48  &  70.12$\pm$0.86  &  82.16$\pm$0.80   \\
         \bottomrule
    \end{tabular}}
\caption{Performance of attention-based aggregation on the node classification task.}
    \label{tab:aggregationatt}
\end{table}

\subsection{Performance on Larger Graph Datasets}
In this section, to underscore our model's capacity for handling large graph datasets, we conduct a series of experiments on the Ogbn datasets obtained from \url{https://ogb.stanford.edu/docs/nodeprop/}, in compliance with the experimental setup detailed in \cite{hu2021ogbdataset}. The corresponding results are encapsulated in \cref{tab:larger}. For the training of our model on the Ogbn-products dataset, we employ a straightforward neighborhood sampling approach, as introduced in GraphSage \cite{hamilton2017inductive}. The GCN results are extracted directly from the leaderboard available at \url{https://ogb.stanford.edu/docs/leader_nodeprop/}. 
We observe that for such large datasets, our model still performs efficiently compared to other methods.

\begin{table}[!htb]
    \centering
 \resizebox{0.3\textwidth}{!}{\begin{tabular}{c|c|ccccc}
    \hline
        Model & Ogbn-Arxiv  & Ogbn-Products \\
        \hline
         HAMGNN &  71.70$\pm$0.27 & 79.87$\pm$0.04  \\
         GCN  & 71.74$\pm$0.29   & 75.64$\pm$0.21  \\
         
         \hline
    \end{tabular}}
    \caption{Node classification results(\%) on Ogbn datasets}
    \label{tab:larger}
\end{table}

\subsection{Computational Cost}
To provide further information, we present the memory usage along with the training and inference time in \cref{tab:modelsize}. When considering efficiency, HamGNN exhibits a reduced training and inference duration in comparison to both HGCN and GIL. Furthermore, it boasts a smaller model size relative to GIL, while maintaining a similar size to GCN/HGCN. 
Overall, the table suggests that HamGNN models require higher computational resources than GCN but are more efficient than HGCN and GIL, and can achieve higher accuracy on low hyperbolicity graph learning tasks.

\begin{table}[!htp]\small
    \centering
    \resizebox{0.5\textwidth}{!}{
    \begin{tabular}{c|c|c|c|c}
    \toprule
         & HamGNN(20) & GCN & HGCN & GIL \\
    \midrule
        Num of para &  113025 & 92231 & 92231 & 189740 \\
        Model Size (MB) & 0.431  & 0.352 & 0.352 & 0.724 \\
        Inference Time (ms) & 3.245  & 1.662 & 4.355 & 14.332 \\
        Training Time (ms) & 9.612  & 6.064 & 38.766 & 60.022\\
    \bottomrule
        
    \end{tabular}}
    \caption{Model size and computation time. All models are using one hidden layer with a hidden dimension of 64 on the CORA dataset for a fair comparison.}
    \label{tab:modelsize}
\end{table}

\subsection{Examination of learned curvature parameters}
In this section, we provide the visualization of the learned Ricci scalar curvature (\url{https://en.wikipedia.org/wiki/Ricci_curvature}) that is computed from our learnable metric $g$. We use one high $\delta$-hyperbolicity dataset, the Cora dataset, and one low $\delta$-hyperbolicity dataset, the Airport dataset. (Note that higher $\delta$-hyperbolicity means \emph{less} hyperbolic.) 

\textbf{Airport:} We observe that the local curvatures learned on the Airport dataset show various values, ranging from -2000 to +2000. Few embeddings show positive curvature, and most of the embeddings show \textbf{negative or near 0 curvature} at the local geometry. This is expected as hyperbolic GNNs (Chami et al., 2019; Liu et al., 2019) have demonstrated that tree-structured datasets have improved classification accuracy if they are embedded into a hyperbolic space with constant negative curvature rather than the zero curvature Euclidean space. Our model automatically learns to achieve this.
The difference is that in HamGNN with \cref{eq:H_geo}, the local curvature varies from point to point based on learning, and some points may even be embedded with a local positive curvature geometry. The visualized learned local curvatures explain why our model can surpass the baselines on the lower $\delta$-hyperbolicity datasets. 

\textbf{Cora:} For the Cora dataset, from the visualization, we observe that almost all the embeddings are located at locations with \textbf{near zero curvature}, i.e., similar to the Euclidean space. This is consistent with our results shown in the paper that HamGNN has a comparable performance with Euclidean-based GNNs on high $\delta$-hyperbolicity datasets. Our HammGNN successfully learns to embed the nodes in spaces that closely resemble Euclidean spaces.

The above curvature visualization further demonstrates that the curvature in HamGNN can successfully adapt for different structures of different datasets, and this learnable metric leads to all-around good performance.

\begin{figure}[!htb]
    \centering
    \includegraphics[width=0.5\textwidth]{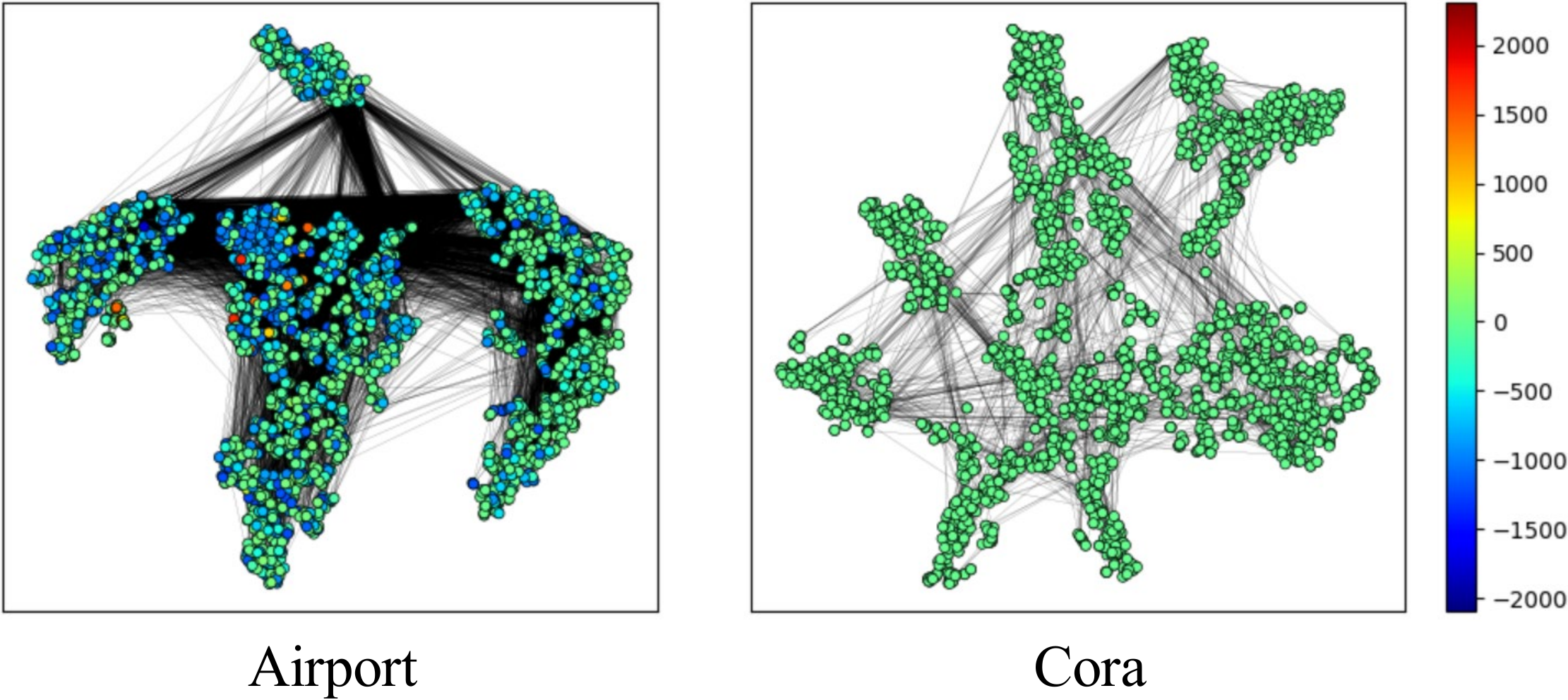}
    \caption{Ricci scalar curvature on the Cora and Airport dataset using t-SNE.}
    \label{fig:tsne}
\end{figure}

\subsection{Over-Smoothing} \label{appsec:app_exp_smooth}
Our proposed model aims to alleviate the over-smoothing issue in GNNs during the feature updating stage rather than the aggregation stage. We are not claiming that HamGNN totally resolves the over-smoothing problem. Over-smoothing typically arises from the repeated application of graph convolution or aggregation operations, which tend to mix and average features from neighboring nodes. However, since our model still employs the traditional aggregation stage, our experiments on over-smoothing also suggest that some of the over-smoothing may originate from the feature updating stage. We continue from \cref{sec:ove_smo} to conduct more experiments on the Cora and Pubmed datasets to demonstrate the resilience of HamGNN against over-smoothing.

From \cref{tab:num_layer_full}, we observe that if the $H_{\mathrm{net}}$  in \cref{eq:H_2} is convex, HamGNN can even retain its classification ability better than vanilla $H_{\mathrm{net}}$ in \cref{eq:H_1}. 
One possible reason has been indicated in \cref{sec:H_2} since now the Hamiltonian formalism degenerates to a Lagrangian formalism with a possible minimization of the dual energy functional \cref{eq:sq}.
In physics, lower energy in most cases indicates a more stable system equilibrium.
Moreover, to further show that with the convex $H_{\mathrm{net}}$  in \cref{eq:H_2} HamGNN can perform better than the vanilla $H_{\mathrm{net}}$ in \cref{eq:H_1} against over-smoothing, we now include more choices of convex network $H_{\mathrm{net}}$ with different  layer sizes and different activation functions (as along as the layer weights from the second layer in $H_{\mathrm{net}}$ are non-negative, and all activation functions in $H_{\mathrm{net}}$ are convex and non-decreasing).
The network details of $H_{\mathrm{net}}$ are given in \cref{tab:hlayers}. The experiment results are shown in \cref{tab:num_layer_full}.
We clearly observe that for different convex functions and on different datasets, HamGNN with convex $H_{\mathrm{net}}$  nearly keeps the full  node classification ability even though we have stacked 20 Hamiltonian layers.

\begin{table*}[!t]
\centering
\makebox[\textwidth][c]{
\begin{tabular}{lcccccc}
\toprule
Dataset  &  Models  &  3 layers   &  5 layers   &  10 layers   &  20 layers  \\
\midrule
\multirow{5}{*}{Cora} 
& GCN &  80.29$\pm$2.29  &  69.87$\pm$1.12  &  26.50$\pm$4.68  &   23.97$\pm$5.42  \\
& HGCN &  78.70$\pm$0.96  &  38.13$\pm$6.20  &  31.90$\pm$0.00  &    26.23$\pm$9.87  \\
& HamGNN \cref{eq:H_1}  &  81.52$\pm$1.27  &  \tb{81.58$\pm$0.73}  &   79.00$\pm$2.17   &   76.20$\pm$0.13\\
& HamGNN \cref{eq:H_2} &  81.84$\pm$0.88  &  81.08$\pm$0.16  &  \tb{81.40$\pm$0.44}  &   \tb{80.58$\pm$0.30}  \\
& HamGNN \cref{eq:H_2} type 2 &  \tb{82.10$\pm$0.80}  &  81.10$\pm$0.10  &  81.06$\pm$1.49  &   79.26$\pm$1.07  \\
\midrule
\multirow{5}{*}{Pubmed} 
& GCN &  77.83$\pm$0.77  &  76.00$\pm$0.87  &   77.53$\pm$1.06  &   56.50$\pm$12.79 \\
& HGCN &  76.38$\pm$0.81  &  77.20$\pm$1.05  &  65.90$\pm$9.44  &   42.16$\pm$2.54  \\
& HamGNN \cref{eq:H_1}  &  78.18$\pm$0.54  &  77.86$\pm$1.45  &  77.73$\pm$1.15   &  76.13$\pm$0.80   \\
& HamGNN \cref{eq:H_2}  &  78.83$\pm$0.46 &  78.43$\pm$0.25  &  78.50$\pm$0.61   &   77.20$\pm$0.69  \\
& HamGNN \cref{eq:H_2} type 2   &  \tb{79.03$\pm$0.58}  &  \tb{78.46$\pm$0.11}  &  \tb{78.53$\pm$0.31}   &   \tb{77.50$\pm$0.44}  \\
\bottomrule
\end{tabular}}
\caption{Node classification accuracy(\%) when increasing the number of layers on the Pubmed dataset.}
\label{tab:num_layer_full}
\end{table*}

\begin{figure*}[!htb]
\centering
\includegraphics[width=1\textwidth]{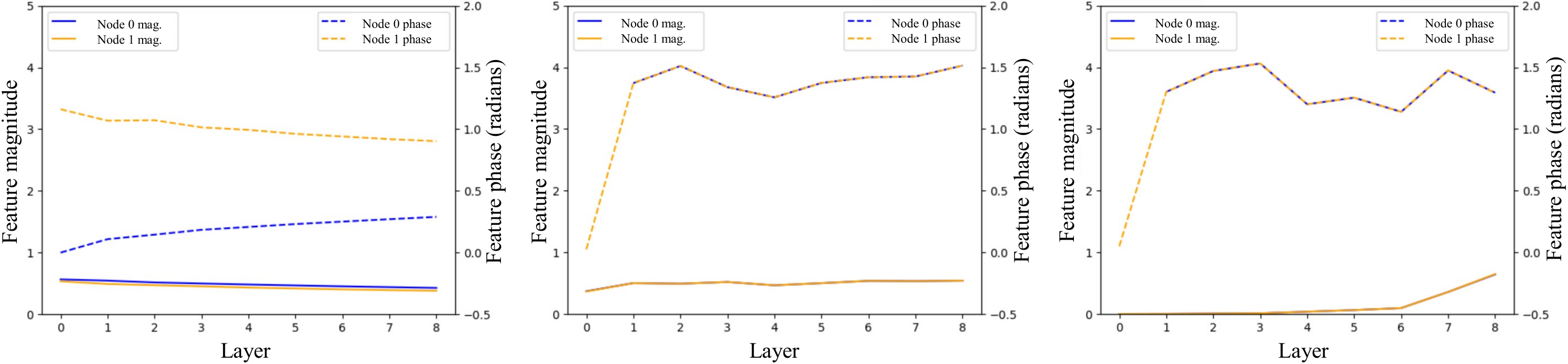}
\caption{Two nodes from different classes and the evolution of their feature vectors over layers. Left: HamGNN, middle: GIL, right: GCN.}
\label{fig:oversmooth}
\end{figure*}

\cref{fig:oversmooth} shows how the node features evolve over 10 layers. We sample a node from the test set of the PubMed dataset and input it to three models, which are 1) HamGNN, 2) GIL, and 3) GCN. Each of these three GNNs contains 10 layers, and we compute the node feature magnitude, which is defined to be the $L_2$-norm of the feature vector, and the node feature phase, which is defined to be the cosine similarity between the output feature at the current layer and the input feature to the first layer.  We can observe from \cref{fig:oversmooth} that HamGNN has its learned node features that change steadily and slowly, while the node features learned by GIL and GCN change abruptly, especially for the feature phases. The features from two nodes of different classes, learned by GIL and GCN, are converging to each other much faster than HamGNN.

In this paper, we have presented extensive experimental analyses supporting HamGNN's ability to mitigate over-smoothing. Further theoretical analysis of our HamGNN is left for future work.

\subsection{Stability}\label{appsec:stability}
\begin{figure*}[!htb]
\centering
\includegraphics[width=1\textwidth]{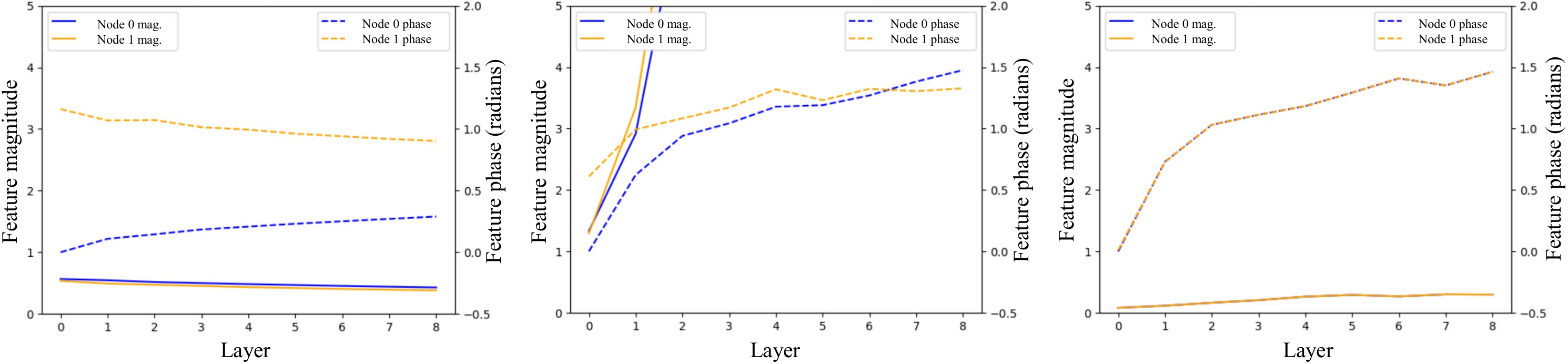}
\caption{Two nodes from different classes and the evolution of their feature vectors over layers. Left: HamGNN, middle: neural ODE using a positive-definite linear layer, right: neural ODE using a negative-definite linear layer.}
\label{fig:stable}
\end{figure*}

A known characteristic of Hamiltonian systems is their energy-conserving nature. Our proposed HamGNN retains this property of stability or conservation. For our analysis, we incorporate three neural ODE-based GNN models: 1) HamGNN, 2) an ODE model employing a positive-definite linear layer as the ODE function, and 3) an ODE model using a negative-definite linear layer as the ODE function. Each of these three GNNs consists of 10 layers.
The results, as presented in \cref{fig:stable}, illustrate noticeable trends: the feature magnitude (analogous to feature energy) learned by HamGNN remains constant across layers, while the feature magnitude learned by the positive-definite model escalates after a few layers and the one learned by the negative-definite ODE model approaches zero. With respect to the feature phase, the phases from two nodes gradually converge when using HamGNN, while those learned by other neural ODEs exhibit abrupt shifts, and the difference between two nodes using the negative-definite ODE is insignificantly small.

The work \cite{haber2017stable} constructs a Hamiltonian-inspired neural ODE and asserts its capability to stabilize gradients, thereby circumventing issues of vanishing and exploding gradients. This resilience to vanishing and exploding gradients may also contribute to the resistance to over-smoothing. The further theoretical investigation remains a subject for future work.

\begin{table*}[!t]
\caption{Neural network modules and the parameters, where $\kappa(x) = 0.5x^2 \text{ if } x>0 \text{ and } = \exp(x)-1 \text{ if } x<0$}
\label{tab:hlayers}
\begin{center}
\newcommand{\tabincell}[2]{\begin{tabular}{@{}#1@{}}#2\end{tabular}}
\begin{tabular}{cccc}
\toprule
\multicolumn{1}{c}{Network}  & {Modules} & {Activation} & {Output Channels} \\ 
\midrule
\cref{eq:H_1}  & \tabincell{c}{Linear \\ tanh \\ Linear }      
& \tabincell{c}{tanh}
& \tabincell{c}{1} \\ 
\midrule

\cref{eq:H_2}  & \tabincell{c}{ReHU\cite{amos2017input} \\ Linear \\ ReHU \\ Linear \\ ReHU \\ Linear \\ ReHU}      
& \tabincell{c}{ReHU}
& \tabincell{c}{1} \\ 
\midrule

\cref{eq:H_2} type 2 & \tabincell{c}{$\kappa$ \\ Linear \\ $\kappa$ \\ Linear \\ $\kappa$ \\ Linear \\ $\kappa$}      
& \tabincell{c}{ReHU}
& \tabincell{c}{1} \\ 
\midrule

\cref{eq:H_3}  & \tabincell{c}{Linear \\ act1 \\ Linear \\ act1}      
& \tabincell{c}{act1}
& \tabincell{c}{576} \\ 
\midrule

\cref{eq:H_4} & \tabincell{c}{Linear \\ tanh \\ Linear }      
& \tabincell{c}{tanh}
& \tabincell{c}{1} \\ 
\midrule

\cref{eq:H_6} & \tabincell{c}{Linear \\ tanh \\ Linear \\ Sigmoid}      
& \tabincell{c}{tanh \\ Sigmoid}
& \tabincell{c}{64} \\ 
\midrule

\cref{eq:H_8} & \tabincell{c}{Linear \\ sin \\ Linear \\ sin}      
& \tabincell{c}{sin}
& \tabincell{c}{H:1 \\ W:128} \\ 
\midrule
$Q_{\mathrm{net}}$ & Linear & identity  & 64 \\
Raw feature compressing FC & Linear  & identity   & 64 \\
\midrule
MLP in \cref{sec:exp} & \tabincell{c}{Linear \\ ReLU \\ Linear \\ ReLU \\ Linear}      
& \tabincell{c}{ReLU}
& \tabincell{c}{Number of classes} \\ 
\bottomrule
\end{tabular}
\end{center}
\end{table*}

\section{Differential Geometry and Hamiltonian System}\label{appsec:app_diff}
In this supplementary material, we review some concepts from a differential geometry perspective. We hope that this overview makes the paper more accessible for readers from the graph learning community.
\subsection{Manifold, Bundles, and Fields}
Roughly speaking, a manifold is a topological space that \emph{locally} looks like Euclidean space. More strictly speaking, a topological space $(M,\calO)$, where $\calO$ is the collection of open sets on space $M$, is called a $d$-dimensional manifold if for every point $x\in M$, we can find an open neighborhood $U\in \calO$ for $x$ and a coordinate map 
\begin{align*}
    q : U \to q(U) \se \Real^d
\end{align*}
that is a \emph{homeomorphism}, where $\Real^d$ is the $d$-dimensional Euclidean space with the standard topology. 
$(U, q)$ is called a \emph{chart} of the manifold, which gives us a numerical representation for a local area in $M$. In this work, we only consider smooth manifolds \cite{lee2013smooth} that  any two overlapped charts are smoothly compatible. The set of all smooth functions from $M$ to $\R$ is denoted as $\smt{M}$.
On top of the smooth manifolds, we can define other related manifolds like the tangent or cotangent bundles and the more general tensor bundles. From bundles, we can define the vector or covector fields and the more general tensor fields. In this work, we mainly consider the manifold with the $2$-forms that are special $(0,2)$ smooth tensor fields with antisymmetric constraints. More specifically, we mainly consider the \emph{symplectic form} \cite{lee2013smooth} with some other light shed on the metric tensor which is another type of $(0,2)$ smooth tensor field. 
\begin{Definition}[tangent vector and tangent space]
   Let $\gamma\cl I \to M$ be a smooth curve through $x\in M$ s.t. $\gamma(0)=x$ and $I$ is an interval neighborhood of $0$. The  \emph{tangent vector} is a directional derivative operator at $x$ along $\gamma$ that is the \emph{linear map}
\begin{align*}
    v_{\gamma,x}\cl  \smt{M} & \xrightarrow{\sim}  \R\\
 f  & \mapsto  (f\circ\gamma)'(0).
\end{align*}
We also call the directional derivative operator as ``velocity'' of $\gamma$ at $0$ and denote it as $\dot{\gamma}(0)$. (More generally, the `velocity'' of $\gamma$ at $t$ is denoted at $\dot{\gamma}(t)$ which is a tangent vector at point $\gamma(t)
\in M$ from the curve reparametrization trick \cite{fecko2006differential}). Correspondingly, the \emph{tangent space} to $M$ at $x$ is the \emph{vector space} over $\R$ with the underlying set
\begin{align}
T_xM \coloneqq \{v_{\gamma,x}\mid \gamma \text{ is a smooth curve and }  \gamma(0)=x\}.
\end{align}
Note for $f\in \smt{M}$, using the chart $(U,q)$ with $x\in U$, we have the local representation
\begin{align*} 
    \begin{split}
        v_{\gamma,x}(f)& \coloneqq (f\circ \gamma)'(0) \\
        & = (f\circ q^{-1}\circ q \circ \gamma)'(0) \\
        & = (q^i\circ \gamma)'(0) \cdot \p_i \big( f\circ q^{-1}\big)\big|_{q(x)},
    \end{split}
\end{align*}
where $q^i$ is the $i$-th component of $q$, $\circ$ is the function composition and $(\cdot)|_{q(x)}$ means evaluating $(\cdot)$ at $q(x)$. Therefore for a local chart around $x$, we have  a basis of $T_xM$ as
    \begin{align*} 
        \bigg\{ \bigg(\frac{\p}{\p q^1}\bigg)_x, ... , \bigg(\frac{\p}{\p q^d}\bigg)_x\bigg\}
    \end{align*} 
and we call it the \emph{chart induced basis}, where  $\bigg(\frac{\p}{\p q^1}\bigg)_x\coloneqq \p_i (f\circ q^{-1})|_{q(x)}$.
\end{Definition}

\begin{Definition}[tangent bundle]
     Given a smooth manifold $M$, the \emph{tangent bundle} of $M$ is the \emph{disjoint union} of all the tangent spaces to $M$, i.e., we have 
\begin{align}
TM \coloneqq\coprod_{x\in M}T_xM,
\end{align}
equipped with the \emph{canonical projection map}
\bi{rrCl}
\tilde{\pi} \cl & TM & \to & M\\
& X & \mapsto & \tilde{\pi}(X),
\ei
where $\tilde{\pi}(X)$ sends each vector in $T_xM$ to the point $x$ at which it is tangent, and $\coprod$ is the disjoint union. Furthermore, the tangent bundle\footnote{In this paper, we may just use the word ``bundle'' to indicate the total space in the bundle.} is a $2d$-dimensional manifold. If $X\in\tilde{\pi}^{-1}(U)\se TM$ with a local chart $(U,q)$ on $M$ s.t. $x\in U$, then $X\in T_{\tilde{\pi}(X)}M$ from the definition. Since $\tilde{\pi}(X)\in U$, $X$ can be written in terms of the \emph{chart induced basis}:
\begin{align}
X = \tilde{p}^i(X) \tvb{q}{i}{\tilde{\pi}(X)},
\end{align}
where $\tilde{p}^1,\ldots,\tilde{p}^{d}$ are smooth scalar functions. We can then define the following map as a local chart for the manifold $T M$ induced from the chart $(U,q)$ on $M$:
\bi{rrCl}
\xi \cl &\tilde{\pi}^{-1}(U) & \to & q(U) \times \R^{d} \se \R^{2d}\\
& X & \mapsto & (q^1(\tilde{\pi}(X)),\ldots,q^d(\tilde{\pi}(X)), \tilde{p}^1(X),\ldots,\tilde{p}^{d}(X)),
\ei
and the topological structure on $TM$ is derived from the initial topology to ensure continuity.
\begin{figure}
    \centering
\tikzset{
pattern size/.store in=\mcSize, 
pattern size = 5pt,
pattern thickness/.store in=\mcThickness, 
pattern thickness = 0.3pt,
pattern radius/.store in=\mcRadius, 
pattern radius = 1pt}
\makeatletter
\pgfutil@ifundefined{pgf@pattern@name@_trsitdias}{
\pgfdeclarepatternformonly[\mcThickness,\mcSize]{_trsitdias}
{\pgfqpoint{0pt}{-\mcThickness}}
{\pgfpoint{\mcSize}{\mcSize}}
{\pgfpoint{\mcSize}{\mcSize}}
{
\pgfsetcolor{\tikz@pattern@color}
\pgfsetlinewidth{\mcThickness}
\pgfpathmoveto{\pgfqpoint{0pt}{\mcSize}}
\pgfpathlineto{\pgfpoint{\mcSize+\mcThickness}{-\mcThickness}}
\pgfusepath{stroke}
}}
\makeatother
\tikzset{every picture/.style={line width=0.75pt}} 

\begin{tikzpicture}[x=0.75pt,y=0.75pt,yscale=-1,xscale=1]

\draw    (234.5,181) .. controls (274.5,151) and (352,161) .. (383.5,181) ;
\draw    (331,105) -- (315,207) ;
\draw    (288,103) -- (287.5,205.5) ;
\draw    (249.5,105.5) -- (269,209.5) ;
\draw    (261.5,168) ;
\draw [shift={(261.5,168)}, rotate = 0] [color={rgb, 255:red, 0; green, 0; blue, 0 }  ][fill={rgb, 255:red, 0; green, 0; blue, 0 }  ][line width=0.75]      (0, 0) circle [x radius= 1.34, y radius= 1.34]   ;
\draw    (288,162.5) ;
\draw [shift={(288,162.5)}, rotate = 0] [color={rgb, 255:red, 0; green, 0; blue, 0 }  ][fill={rgb, 255:red, 0; green, 0; blue, 0 }  ][line width=0.75]      (0, 0) circle [x radius= 1.34, y radius= 1.34]   ;
\draw    (321.5,163) ;
\draw [shift={(321.5,163)}, rotate = 0] [color={rgb, 255:red, 0; green, 0; blue, 0 }  ][fill={rgb, 255:red, 0; green, 0; blue, 0 }  ][line width=0.75]      (0, 0) circle [x radius= 1.34, y radius= 1.34]   ;
\draw [color={rgb, 255:red, 242; green, 6; blue, 6 }  ,draw opacity=1 ]   (321.5,163) -- (327.16,130.47) ;
\draw [shift={(327.5,128.5)}, rotate = 99.87] [color={rgb, 255:red, 242; green, 6; blue, 6 }  ,draw opacity=1 ][line width=0.75]    (4.37,-1.32) .. controls (2.78,-0.56) and (1.32,-0.12) .. (0,0) .. controls (1.32,0.12) and (2.78,0.56) .. (4.37,1.32)   ;
\draw  [draw opacity=0][pattern=_trsitdias,pattern size=6pt,pattern thickness=0.75pt,pattern radius=0pt, pattern color={rgb, 255:red, 184; green, 233; blue, 134}][dash pattern={on 4.5pt off 4.5pt}] (314,87.5) -- (379.5,88.5) -- (346.5,215) -- (305,211.5) -- (310,148) -- cycle ;
\draw [color={rgb, 255:red, 144; green, 19; blue, 254 }  ,draw opacity=1 ]   (327.5,128.5) .. controls (315.8,136.79) and (307.43,141.75) .. (320.46,161.45) ;
\draw [shift={(321.5,163)}, rotate = 235.38] [color={rgb, 255:red, 144; green, 19; blue, 254 }  ,draw opacity=1 ][line width=0.75]    (4.37,-1.32) .. controls (2.78,-0.56) and (1.32,-0.12) .. (0,0) .. controls (1.32,0.12) and (2.78,0.56) .. (4.37,1.32)   ;
\draw [color={rgb, 255:red, 5; green, 206; blue, 249 }  ,draw opacity=1 ]   (310,162) .. controls (328,164) and (332.5,163.5) .. (358.5,169.5) ;

\draw (323.5,166.4) node [anchor=north west][inner sep=0.75pt]  [font=\tiny]  {$x$};
\draw (323.2,94.2) node [anchor=north west][inner sep=0.75pt]  [font=\tiny]  {$T_{x} M$};
\draw (333,127.4) node [anchor=north west][inner sep=0.75pt]  [font=\tiny,color={rgb, 255:red, 247; green, 5; blue, 5 }  ,opacity=1 ]  {$X\in T_{x} M$};
\draw (344,169.4) node [anchor=north west][inner sep=0.75pt]  [font=\tiny,color={rgb, 255:red, 5; green, 238; blue, 248 }  ,opacity=1 ]  {$U$};
\draw (305.5,139.4) node [anchor=north west][inner sep=0.75pt]  [font=\tiny,color={rgb, 255:red, 144; green, 19; blue, 254 }  ,opacity=1 ]  {$\tilde{\pi}$};
\draw (359.5,104.9) node [anchor=north west][inner sep=0.75pt]  [font=\tiny,color={rgb, 255:red, 74; green, 134; blue, 14 }  ,opacity=0.98 ]  {$\tilde{\pi}^{-1}(U)$};
\draw (382.5,174.9) node [anchor=north west][inner sep=0.75pt]  [font=\tiny]  {$M$};

\end{tikzpicture}
 \caption{Visualization of a 1-dimensional manifold and its tangent bundle.}
    \label{appfig:my_label}
\end{figure}
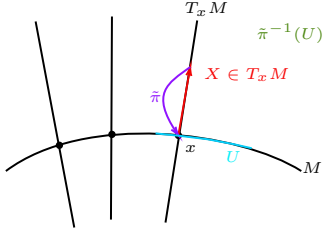

\end{Definition}

\begin{Definition}[vector field]
     A vector field on $M$ is a smooth section \cite{lee2013smooth} of the tangent bundle, i.e.\ a smooth map $\sigma\cl M \to TM$ such that $\tilde{\pi} \circ \sigma = \id_M$, where $\id_M$ is the identity map on $M$.
\begin{align}
\begin{tikzcd}
TM \ar[dd,shift left,"\tilde{\pi}"] \\
\\
M \ar[uu,shift left,"\sigma"]
\end{tikzcd}
\end{align}
 We denote the set of all vector fields on $M$ by $\Gamma(TM)$:
\begin{align}
\Gamma(TM) \coloneqq \{\sigma \cl M \to TM \mid \sigma \text{ is smooth and }\tilde{\pi}\circ\sigma=\id_M\}.
\end{align}
\end{Definition}

\begin{Definition}[cotangent vector, cotangent bundle, dual basis, and covector field]
For the vector space $T_xM$, a continuous linear functional from $T_xM$ to $\R$ is called a cotangent vector at $x$. The set of all such linear maps is denoted as $T_x^*M$ which is the dual vector space of $T_xM$.  For  $f\in \smt{M}$, at each point $x$, we define the following linear operator in  $T_x^*M$
\bi{rrCl}
(\d f)_x\cl & T_xM& \to & \R \\
& X_x & \mapsto & (\d f)_x(X_x) \coloneqq X_x(f).
\ei
Given a chart $(U,q)$ with $x\in U$ and its \emph{chart induced basis}, the \emph{dual basis} for the dual space $T_x^*M$ is the set
 \begin{align*}
     \left\{\left(\d q^1\right)_x,\ldots,\left(\d q^d\right)_x\right\},
 \end{align*}
    where we have $(\d q^a)_x \left( \tvb{q}{b}{x} \right) =  \tvb{q}{b}{x} \left(q^a\right) =\delta^a_b$ with $\delta^a_b=1$ iff $a=b$ and $\delta^a_b=0$ otherwise. We call it the \emph{chart  induced dual basis }. Analogous to the above definition of the vector field, we can define the cotangent bundle of $M$ as 
    \begin{align}
T^*M \coloneqq \coprod_{x\in M} T^*_xM
\end{align}
which is again a $2d$-dimensional manifold equipped with the \emph{canonical projection map}
\begin{align}
\begin{aligned}
    {\pi} \cl  T^*M & \to  M\\
 \omega & \mapsto  \pi(\omega), \label{appeq:co_pro}
\end{aligned}
\end{align}
where $\pi(\omega)$ sends each vector in $T^*_xM$ to the point $x$ at which it is cotangent.  If $\omega\in{{\pi}}^{-1}(U)\se T^*M$ with a local chart $(U,q)$ s.t. $x\in U$, then $\omega\in T^*_{\pi(\omega)}M$ from the definition. Since $\pi(\omega)\in U$, $\omega$ can be written in terms of the \emph{chart induced dual basis}:
\begin{align}
\omega = {p}_i(\omega) (\d q^i)_{{\pi}(\omega)},
\end{align}
where ${p}_1,\ldots,{p}_{d}$ are smooth scalar functions. We can then define the following map as a local chart for manifold $T^*M$ induced from the chart $(U,q)$ on $M$:
\bi{rrCl}
\xi \cl &\pi^{-1}(U) & \to & q(U) \times \R^{d} \se \R^{2d}\\
& \omega & \mapsto & (q^1(\pi(\omega)),\ldots,q^d(\pi(\omega)), {p}_1(\omega),\ldots,{p}_{d}(\omega)),
\ei
and the topological structure on $T^*M$ is derived from the initial topology to ensure continuity.
The covector fields are smooth sections of $T^*M$. The set of all covector fields is denoted as $\Gamma(T^*M)$. 
\end{Definition}

\begin{Definition}[tensor field and $2$-form]
The tensor field can be defined using the smooth sections on tensor bundles analogously to the vector fields or the covector fields. We refer readers to \cite{lee2013smooth} for more details. Here, instead of a rigorous definition, we show some basic properties of the tensor fields. For a  $(r,s)$ tensor field $\tau$, at $x\in M$, it is a multilinear map
\begin{align*}
    \tau_x\cl \underbrace{T_x^*M\times \cdots \times T_x^*M}_{r \text{ copies}} \times \underbrace{T_xM \times \cdots \times T_xM}_{s \text{ copies}} \to \R.
\end{align*}
The differential $k$-form $\omega$ is the $(0,k)$ tensor field that admits  alternating \cite{lee2013smooth}. Specifically, for the $2$-form $\omega$, at each point $x$, $\omega_x$ is a \emph{antisymmetric} $(0,2)$ tensor 
\begin{align}
&\omega_x\cl T_xM \times T_xM \to \R \ \\
&\st \omega_x(X_1, X_2) = - \omega_x(X_2,X_1) \ \forall X_1,X_2\in  T_xM
\end{align}
which in other words, $\omega$ satisfies
\begin{align}
&\omega\cl \Gamma(TM) \times \Gamma(TM) \to \smt{M} \ \\
&\st \omega(X, Y) = - \omega(Y,X) \ \forall X,Y\in  \Gamma(TM) \label{appeq:dww=0}
\end{align}
In local chart representation, we have that every $k$-form $\omega$ can be expressed \emph{locally} on $U$ as
\begin{align}
\omega = \omega_{a_1\cdots a_k} \, \d x^{a_1} \wedge \cdots \wedge \d x^{a_k},
\end{align}
where $\omega_{a_1\cdots a_k}\in \mathcal{C}^\infty(U)$, $1\leq a_1 < \cdots < a_k \leq \dim M$ are increasing sequences and $\d x^{a_1} \wedge \cdots \wedge \d x^{a_k}$ is the wedge product \cite{lee2013smooth}. Here we could abstractly view the set $\{\d x^{a_1} \wedge \cdots \wedge \d x^{a_k}\}_{a_1,\ldots,a_k}$, with $a_i$ enumerated from $1$ to $d$,  abstractly as a basis without more illustrations of the wedge product (We refer readers to \cite{lee2013smooth} for more details).
\end{Definition}    

\begin{Definition}[integral curve]\label{appdef:int_cur}
Given  a vector field $X$ on $M$, an integral curve of $X$ is a differentiable curve $\gamma: I \rightarrow M$, where $I\se \R$ is an interval, whose velocity at each point is equal to the value of $X$ at that point:
\begin{align}
\dot{\gamma}(t)=X_{\gamma(t)} \quad \text { for all } t \in I .
\end{align}
If $0 \in J$, the point $\gamma(0)$ is called the starting point of $\gamma$.  From Picard’s theorem \cite{hartman2002ordinary}, we know that locally we always have an interval $I$ on which the solution exists and is necessarily unique.
\end{Definition}

\begin{Definition}[exterior derivative and closed form]
    The exterior derivative is a linear operator that maps $k$-forms to $k+1$-forms. In local chart representation, we have that  if $\omega$ is a $k$-form on $M$ with the 
    local representation as
\begin{align}
\omega = \omega_{a_1\cdots a_k}\d x^{a_1}\wedge \cdots \wedge \d x^{a_k}.
\end{align}
Then, we have the exterior derivative
\begin{align}
   \d \omega & =  \d \omega_{a_1\cdots a_k}\wedge\d x^{a_1}\wedge \cdots \wedge \d x^{a_k}\nn
 & =  \partial_b \omega_{a_1\cdots a_k}\d x^b\wedge\d x^{a_1}\wedge \cdots \wedge \d x^{a_k}. \label{appeq:ext_der}
\end{align}
A form $\omega$ is called \emph{closed} if $\d \omega = 0$. 
\end{Definition}
\begin{Theorem}
    From \cite{rudin1976principles,lee2013smooth}, we know that 
\begin{align}
    \d\circ \d \equiv 0, \label{appeq:dd}
\end{align} which is a dual statement that the boundary of the boundary of a manifold is empty from Stokes'~theorem.
\end{Theorem}
\subsection{Hamiltonian Vector Fields on Symplectic Cotangent Bundle}
\begin{Definition}[symplectic vs. Riemannian] Let $M$ be a smooth manifold. 
\begin{enumerate}
    \item symplectic form: A \emph{$2$-form} (so it is antisymmetric) $\omega$ is said to be a \emph{symplectic form} on $M$ if  it is closed, i.e,
    $$\d \omega = 0,$$
    and it is \emph{non-degenerate}, i.e,
\begin{align}
(\forall \, Y\in \Gamma(TM) : \omega(X,Y) = 0) \Rightarrow X = 0.
\end{align}
\item  metric tensor: A \emph{$(0,2)$ tensor field} $g$ is said to be a \emph{Riemannian metric} on $M$ if  it is \emph{non-degenerate} and symmetric at each $x$, i.e.,
\begin{align}
&g_x\cl T_xM \times T_xM \to \R \ \\
&\st g_x(X_1, X_2) = g_x(X_2,X_1) \ \forall X_1,X_2\in  T_xM
\end{align}

\end{enumerate}
\end{Definition} 

\begin{Remark}
A manifold equipped with a symplectic form  $\omega$ is called a \emph{symplectic manifold}, while a manifold equipped with a metric tensor $g$ is called a \emph{pseudo-Riemannian manifold}. The Riemannian metric $g$ is a $(0,2)$-tensor field measuring the norms of tangent vectors and the angles between them.  To some extent, the ``shape structure'' of the manifold $M$ is only available if we equipped $M$ with a metric $g$.
\begin{itemize}
   \item From the above definition, we know the symplectic form  and the metric tensor are both nondegenerate bilinear $(0,2)$ tensor fields. One difference is the symplectic form is antisymmetric while the metric tensor is symmetric.  At each point $x\in M$, if we use the local chart coordinate representation, the $(0,2)$ tensor can be represented as the following matrix multiplication
\begin{align*}
 \tilde{   p}^{\T} W_x\tilde{p} 
\end{align*}
   where $d\times d$ matrix $W$ is symmetric for  metric tensor and is skew-symmetric for symplectic form. $\tilde{p}^{\T}$ is the transpose of the velocity representation (which is a numerical vector) in a local chart.
    \item Because of \emph{non-degenerate}, on a symplectic manifold $M$, we can define an isomorphism between $\Gamma(T M)$ and $\Gamma(T^* M)$ by mapping a vector field $X \in \Gamma(T M)$ to a 1-form $\eta_V \in \Gamma(T^* M)$, where
\begin{align}
\eta_X(\cdot)\coloneqq \omega^2(\cdot, X)
\end{align}
Similarly, on a  pseudo-Riemannian manifold,  we can define an isomorphism between $\Gamma(T M)$ and $\Gamma(T^* M)$ by mapping a vector field $X \in \Gamma (T M)$ to a 1-form $\in \Gamma(T^* M)$, where
\begin{align}
\alpha_g: T M & \longrightarrow T^* M.
\end{align}
In a local chart coordinate representation, $\alpha_g=g_{i j}$ and its inverse $\alpha_g^{-1}=g^{i j}$ with $\sum_{j=1}^m g_{ij} g^{jk}=\delta_i^k$ and $\delta_i^k=1$ iff $i=k$ and $0$ otherwise. Note the components of the metric and the inverse metric are all taken in a given chart without explicitly mentioning them.
\end{itemize}
\end{Remark}

\begin{Definition}[Hamiltonian flow and orbit]\label{appdef:ham_orb}
     For a general symplectic manifold $M$ with a symplectic form $\omega^2$, if we have  $H\in \smt{M}$, then $\d H$ is a differential $1$-form on $M$.  We define the vector field called the \tb{Hamiltonian flow $X_H$ associated to the Hamiltonian $H$}, which satisfies that
\begin{align*}
\eta_{X_H}(\cdot)=\d H(\cdot).
\end{align*}
The integral curves of are called \tb{Hamiltonian orbits of $H$}:
\begin{align}
\dot{\gamma}(t)=\left(X_{H}\right)_{\gamma(t)} \quad \text { for all } t \in I . \label{appeq:ham_orb}
\end{align}
where $\left(X_{H}\right)_{\gamma(t)}$ is the tangent vector at $\gamma(t)\in M$.
\end{Definition}

\begin{Definition}[Poincar\'e $1$-form and $2$-form]
    On the cotangent bundle $T^*M$ of a manifold $M$, we have a natural symplectic form, called the \emph{Poincar\'e $1$-form} 
    \begin{align}
     \theta_{\mathrm{Poincar\acute{e}}}^1 = p_i \d q^i.
\end{align}
Therefore, by the exterior derivative, we have the \emph{Poincar\'e $2$-form} 
\begin{align}
    \omega_{\mathrm{Poincar\acute{e}}}^2 = \d \theta^1 = \d(p_i \d q^i) = \sum_i \d p_i \wedge \d q^i
\end{align}
which is closed from \cref{appeq:dd}. Therefore \emph{Poincar\'e $2$-form} is a  symplectic form on the cotangent bundle and cotangent bundles are the natural \tb{phase spaces} of classical mechanics \cite{de2011generalized}.
\end{Definition}

For more general symplectic forms on the cotangent bundle, we can again use \cref{appeq:dd} to construct the closed $2$-form from a $1$-form which is potentially symplectic:
\begin{Corollary}[\cite{ChenNeurIPS2021}] \label{appcor:dd}
      According to \cref{appeq:dd}, on the cotangent bundle $T^*M$ of a manifold $M$, from a $1$-form,
    \begin{align*}
     \theta^1 = f_i \d q^i,
\end{align*}
we derive a closed $2$-form using the exterior derivative \cref{appeq:ext_der}, and its local representation is given by the following 
\begin{align*}
    \omega^2 = \d \theta^1 = \d(f_i \d q^i) = \sum_{i< j}  \left(\p_if_j -\p_jf_i \right)\d p_i \wedge \d q^j
\end{align*}
\end{Corollary}
\begin{Remark}
Note, strictly speaking, we only can get the necessary  ``closed'' condition for $\omega^2$ to be potentially symplectic. However, it is enough for our use in our proposed framework. 
\end{Remark}

We now state one of the most fundamental results in symplectic geometry that links the general symplectic form to the special Poincar\'e $2$-form.
\begin{Theorem}[Darboux \cite{lee2013smooth}]\label{the:Darboux}
    Let $(\tilde{M}, \tilde{\omega}^2)$ be a $2d$-dimensional symplectic manifold. For any $x \in \tilde{M}$, there are smooth coordinates $\left(\tilde{q}^1, \ldots, \tilde{q}^d, \tilde{p}^1, \ldots, \tilde{p}^n\right)$ centered at $x$ in which $\omega^2$ has the coordinate representation
\begin{align}
\tilde{\omega}^2=\sum_{i=1}^n \d \tilde{q}^i \wedge \d \tilde{p}^i.\label{eq:backto_po}
\end{align}
as a Poincar\'e $2$-form, any coordinates satisfying \cref{eq:backto_po} are called Darboux or symplectic coordinates.
\end{Theorem}
\begin{Remark}\label{rem:Darboux}
Therefore, for any symplectic form $w^2$ on the above cotangent bundle $T^*M$ of $M$, we can always find a symplectic coordinate with the \emph{Poincar\'e $2$-form}.
\end{Remark}

\begin{Corollary}[\cite{da2008lectures}]\label{appcor:energy_per}
      According to \cref{appeq:dww=0}, from the antisymmetric of the $2$-forms, we have 
    \begin{align}
     \omega^2(X_H,X_H)=0
\end{align}
which implies that hamiltonian vector fields preserve their hamiltonian functions $H$.
\end{Corollary}
\begin{Remark}
In physics, hamiltonian functions are typically energy functions for a physical system. \cref{appcor:energy_per} indicates if the system updates its status according to the hamiltonian vector fields, the  time-evolution of the system follows the law of conservation of energy.
\end{Remark}

\begin{Definition}[Hamiltonian orbit generated from Poincar\'e $2$-form \cite{lee2013smooth}]\label{appdef:ham_orb_poin}
    The Hamiltonian orbit generated by the Hamiltonian flow $X_H$ on the  cotangent bundle equipped with the Poincar\'e $2$-form given in local coordinates $(q, p)$ is given by
\begin{align}
\dot{q}^i =\frac{\partial H}{\partial p_i}, \quad
\dot{p}_i =-\frac{\partial H}{\partial q^i}.\label{appeq:poincare}
\end{align}
\end{Definition}

\begin{Definition}[cogeodesic orbits]
    If additionally $M$ is equipped with a metric tensor $g$, i.e, if $M$ is a pseudo-Riemannian manifold with metric $g$, and if we set the hamiltonian $H$ on $T^*M$ as
    \begin{align*}
        H(q, p)=\frac{1}{2} g^{i j}(q) p_i p_j,
    \end{align*}
     the Hamiltonian orbit generated from Poincar\'e 2-form is given by
\begin{align}
\dot{q}^i =\frac{\partial H}{\partial p_i}=g^{i j} p_j, \quad
\dot{p}_i =-\frac{\partial H}{\partial q^i}=-\frac{1}{2} \p_i g^{j k} p_j p_k. \label{appeq:geo}
\end{align}
It is called the \tb{cogeodesic orbits of $(M,g)$.}  The canonical projection of cogeodesic orbits under $\pi$ \cref{appeq:co_pro} is called \emph{geodesic} on the base manifold $M$ which generalizes the notion of a "straight or shortest line" to manifold where the length is measured by the metric tensor.
\end{Definition}

\end{document}
